\documentclass[10pt,twocolumn,letterpaper]{article}

\usepackage{cvpr}              

\definecolor{cvprblue}{rgb}{0.21,0.49,0.74}
\usepackage[pagebackref,breaklinks,colorlinks,allcolors=cvprblue]{hyperref}

\usepackage{amsfonts}       
\usepackage{nicefrac}       
\usepackage{microtype}       
\usepackage{amsmath}
\usepackage{amssymb}
\usepackage{amsthm}
\usepackage{algorithm}
\usepackage{algpseudocode}
\usepackage{float}
 \usepackage{gensymb}
\usepackage{multirow}
 
\usepackage{graphicx}
\usepackage{booktabs}
\usepackage{tabularx}
\usepackage{enumitem}
\usepackage{textcomp}
\usepackage{mathtools}
\usepackage{pgfplots}
\usepackage{colortbl}  
\usepackage{xcolor}     
\usepackage{bbm}

\usepackage[accsupp]{axessibility}  

\newcommand{\graytext}[1]{\textcolor{gray}{ #1}}

\def\paperID{10182} 
\def\confName{CVPR}
\def\confYear{2025}

\title{PatchGuard: Adversarially Robust Anomaly Detection and Localization through Vision Transformers and Pseudo Anomalies}

\author{
Mojtaba Nafez$^{1}$ \quad Amirhossein Koochakian$^{1}$ \thanks{Equal Contribution} \quad Arad Maleki$^{1}$ \footnotemark[1] \\
Jafar Habibi$^1$ \quad Mohammad Hossein Rohban$^1$\\
$^1$Sharif University of Technology, Iran \\
{\footnotesize \texttt{\{mojtaba.nafez77, amir.kochakian12, arad.maleki02, rohban\}@sharif.edu}} 
}

\begin{document}
\maketitle 
\begin{abstract}

Anomaly Detection (AD) and Anomaly Localization (AL) are crucial in fields that demand high reliability, such as medical imaging and industrial monitoring. However, current AD and AL approaches are often susceptible to adversarial attacks due to limitations in training data, which typically include only normal, unlabeled samples. This study introduces PatchGuard, an adversarially robust AD and AL method that incorporates pseudo anomalies with localization masks within a Vision Transformer (ViT)-based architecture to address these vulnerabilities.
We begin by examining the essential properties of pseudo anomalies, and follow it by providing theoretical insights into the attention mechanisms required to enhance the adversarial robustness of AD and AL systems. We then present our approach, which leverages Foreground-Aware Pseudo-Anomalies to overcome the deficiencies of previous anomaly-aware methods. Our method incorporates these crafted pseudo-anomaly samples into a ViT-based framework, with adversarial training guided by a novel loss function designed to improve model robustness, as supported by our theoretical analysis.
Experimental results on well-established industrial and medical datasets demonstrate that PatchGuard significantly outperforms previous methods in adversarial settings, achieving performance gains of $53.2\%$ in AD and $68.5\%$ in AL, while also maintaining competitive accuracy in non-adversarial settings. The code repository is available at \href{https://github.com/rohban-lab/PatchGaurd}{here} .

\end{abstract}    
\begin{figure}[ht]
        \centering
        \includegraphics[width=\linewidth]{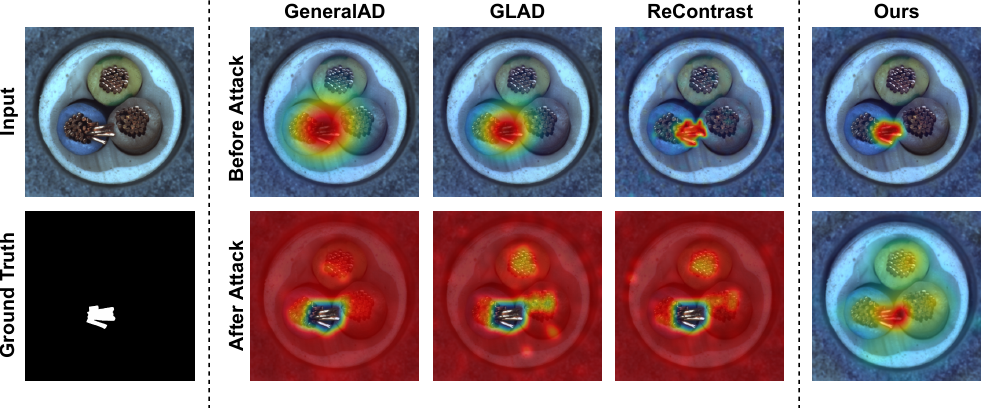}
        \caption{\textbf{Impact of Adversarial Attacks on Anomaly Localization Methods:} Localization maps for multiple methods are shown before and after a PGD-1000 attack, illustrating the vulnerability of existing methods to adversarial     attacks, even when performing perfectly in clean conditions. Our proposed method demonstrates enhanced robustness in these adversarial scenarios.}

        \label{fig:motivation figure}
\end{figure}

\section{Introduction} \label{sec:intro}
In recent years, Anomaly Detection (AD) and Anomaly Localization (AL) have become critical techniques across various fields, including medical imaging, fraud detection, and industrial monitoring \cite{jeffrey2023review, luo2021deep, fernando2021deep, tschuchnig2022anomaly, liu2024deep, yang2024generalized, wang2021anomaly}. While state-of-the-art (SOTA) AD and AL methods exhibit near-perfect performance on standard benchmarks \cite{bergmann2019mvtec, zou2022spot}, they frequently face significant performance degradation when subjected to adversarial attacks, as illustrated in Figure \ref{fig:motivation figure} \cite{strater2024generalad, yao2024glad, zhang2023diffusionad, chen2024unified, mousakhan2023anomaly, tack2020csi}. These attacks involve introducing subtle, imperceptible perturbations into the input data, misleading models to misclassify normal\footnote{In this study, the term `\textit{normal}' refers to inlier samples, not to be mistaken with Gaussian distribution.} samples as anomalies or to overlook actual anomalies \cite{szegedy2013intriguing, goodfellow2014explaining, akhtar2018threat, madry2017towards, zhang2021survey}. To address this vulnerability, a variety of strategies have been proposed to strengthen the adversarial robustness of AD systems \cite{azizmalayeri2022your, lo2022adversarially, chen2020robust, shao2020open, bethune2023robust, goodge2021robustness}. However, despite these advancements, the resilience of these methods remains limited, with even mild adversarial attacks still causing noticeable performance drops, even on widely used anomaly benchmarks \cite{bergmann2019mvtec, zou2022spot}. Furthermore, adversarial robustness in AL remains an unexplored area, highlighting a critical gap in the current research.

Adversarial training, a strategy designed to improve model robustness in classification tasks, incorporates adversarial samples into the training process \cite{madry2017towards, yuan2019adversarial}. Despite its success in standard classification, this approach proves less effective in AD and AL scenarios \cite{mirzaeirodeo}. The primary obstacle to achieving adversarial robustness in AD lies in the absence of anomalous samples during training \cite{chen2020robust, chen2021atom}. Without such samples, models cannot be trained on adversarial perturbations specific to anomalous data \cite{salehi2021unified, bendale2015towards, perera2021one, mirzaeirodeo}, leaving them vulnerable to attacks on perturbed anomalies \cite{chen2020robust, chen2021atom, azizmalayeri2022your}. To address this issue in the AD setup, recent efforts have combined Outlier Exposure (OE) \cite{hendrycks2018deep} with adversarial training \cite{azizmalayeri2022your, chen2020robust, chen2021atom, mirzaei2025mitigatingspuriousnegativepairs}. However, this method struggles when applied to high-resolution, real-world datasets \cite{mirzaeirodeo}. This challenge is even more pronounced in AL settings, as robust AL methods must perform pixel-level classification, which inherently requires pixel-level supervision,  while at the same time operating within the same data constraints as the AD framework \cite{huang2025real, zhang2024gan}. Despite ongoing research, the adversarial robustness of AL methods remains an open problem, with no comprehensive solutions proposed to date.

In this study, we identify two essential properties that OE techniques must satisfy for effective application in adversarially robust AD and AL. First, the outlier exposures must include localization maps to facilitate the training of robust AL models. This requirement aligns with prior research on adversarially robust segmentation models, which underscores the importance of generating localization maps to enhance model robustness \cite{luc2016semantic, lee2022anti, wang2019weakly}. Second, as noted by \cite{mirzaeirodeo}, OE samples should be ``near-distributed" relative to normal data, meaning they should share similar semantic and stylistic attributes. This condition is consistent with the literature on adversarial robustness, which highlights the advantages of near-distribution decision boundaries samples for optimizing the model robustness \cite{xing2022artificially}.

Moreover, the recent success of Vision Transformer (ViT)-based architectures in clean AD and AL tasks \cite{cohen2021transformaly, strater2024generalad, li2024musc, jeong2023winclip}, attributed to their global contextual modeling \cite{dosovitskiy2020image}, motivated us to explore their potential in adversarial settings. For an input image $x$ and a fixed attention layer $\ell$, we define the \textit{Attention Degree} of $\ell$ under $x$ as the average number of input embeddings each output embedding attends to. As shown in Figure \ref{fig:theory figure}, we observe that images that induce higher attention degrees in the model are more resistant to adversarial attacks. Furthermore, adversarial training appears to implicitly steer the model toward achieving a high attention degree. We provide a theoretical analysis and justification for these findings in Section \ref{theory}.

Building on these insights, we propose PatchGuard, an adversarially robust AD and AL method that addresses limitations in current OE-based methods and leverages attention-based insights. To address OE challenges, we generate pseudo-anomaly samples from normal data, ensuring they meet two essential criteria: availability of localization maps and near-distribution alignment with inliers. To create samples that exhibit anomalous characteristics while remaining close to the inlier distribution, we introduce Foreground-Aware Pseudo-Anomaly Generation. This approach identifies causal regions in normal samples using an enhanced Grad-CAM \cite{gradcam} feature attribution technique with tailored augmentations. These identified regions are then selectively distorted, drawing on insights from \cite{noohdani2024decompose}, to produce anomaly samples that retain stylistic and semantic coherence with the original data. By precisely controlling the distortion location, we can generate accurate localization maps for the pseudo-anomalies, effectively meeting both criteria for robust AD and AL.

In the next step, we leverage normal samples and pseudo-anomalies to perform adversarial training on a ViT-based architecture. Building on the provided insights, we introduce a novel loss function designed to increase the model's last-layer attention degree. As our theoretical justifications suggest, this property significantly enhances the model's adversarial robustness in both AD and AL settings.

\textbf{Contribution.} In this paper, we propose PatchGuard, an adversarially robust AD and AL method. To the best of our knowledge, we are the first to raise and effectively address the issue of adversarial robustness in AL. We provide insights into the shortcomings of previous OE-based methods and introduce the Foreground-Aware Pseudo-Anomaly Generation framework to address these limitations. Additionally, we provide insights and theoretical justifications on enhancing the adversarial robustness of ViT-based architectures, and propose a novel loss function aligned with these findings.
We evaluate PatchGuard in both clean and adversarial settings, rigorously testing its robustness under several strong attacks, including PGD-1000 \cite{madry2017towards}, $\text{A}^3$ \cite{liu2022practical}, and SEA \cite{croce2024robust}. Results demonstrate that PatchGuard significantly outperforms existing methods in adversarial contexts, achieving an AUROC improvement of $53.2\%$ in AD and $68.5\%$ in AL, while maintaining competitive performance in standard conditions. Our evaluations span various well-established industrial and medical datasets, including MVTec \cite{bergmann2019mvtec}, VisA \cite{zou2022spot}, and BraTS2021 \cite{baid2021rsna}, highlighting PatchGuard's applicability.

\section{Related Work}
\label{sec:related_work}

From an adversarial robustness perspective, anomaly detection and localization methods can be mainly classified into three categories: 1) methods with anomaly-free training, 2) methods that leverage anomalies at the embedding level, and 3) methods that leverage anomalies at the input level. We experimentally adapted methods from each category to adversarial training processes, and the results reveal they still exhibit significant vulnerabilities even after adversarial training. The first category consists of methods such as Recontrast\cite{guo2024recontrast}, Transformaly \cite{cohen2021transformaly}, and PatchCore \cite{roth2021total}, that operate on the hypothesis that models trained exclusively on normal samples can reconstruct normal images well but fail to do so with abnormal regions. These methods lack robustness, since the model does not encounter any anomaly data during adversarial training, meaning that when adversarial perturbations are added to anomaly data, the model still remains vulnerable. The second category, which includes methods such as SimpleNet \cite{liu2023simplenet}, GeneralAD \cite{strater2024generalad}, UniAD \cite{you2022unified}, and DSR \cite{zavrtanik2022dsrdualsubspace}, utilize a feature extractor during training to obtain feature vectors of the normal set. By perturbing these features, they create anomaly features and primarily train a discriminator over these normal and anomaly features. However, these methods lack robustness due to their reliance on highly vulnerable feature extractors and the absence of anomaly data in the input space, which prevents proper training of a robust feature extractor. The third category consists of DRAEM \cite{zavrtanik2021draem}, NSA \cite{schluter2022natural}, Cutpaste \cite{li2021cutpaste}, and GLASS \cite{chen2024unified}, generate pseudo-anomaly data during training. Nevertheless, some of these methods lack access to masks or do not use a foreground-aware generation process. Despite their astonishing success in clean settings, we empirically show in Table \ref{tab:adapting clean methods table} that even when adversarially trained, they exhibit unsatisfactory results. Furthermore, while anomaly localization has not been explored in an adversarial setup, some other works have pursued anomaly detection in an adversarial setting, including PrincipaLS \cite{lo2022adversarially}, ZARND \cite{mirzaei2024killing}, and Rodeo \cite{mirzaeirodeo}. These approaches have achieved relatively better results by incorporating pseudo-anomaly generation processes and adversarial training. However, their performance falls short in real-world high-resolution datasets (as presented in Table \ref{tab:Main_table}). For more details about previous works, see Appendix \ref{appendix:additional related work}.

\section{Preliminaries}
\label{sec:preliminaries}

\textbf{Anomaly Detection (AD) and Localization (AL):} AD involves identifying instances that deviate from expected patterns, while AL focuses on localizing abnormal subregions within an image. In AD, a model assigns an anomaly score \( S(x; f) \) to each sample \( x \) using model \( f \) ; if this score exceeds a predefined threshold, the sample is classified as anomalous. Conversely, lower scores indicate normal samples. In AL, a model generates an anomaly map  \( M(x; f) \) using model \( f \) , where each pixel of the map corresponds to an anomaly score for the respective pixel in the input image.

\textbf{Adversarial Robustness of Anomaly Detectors:}  
An adversarial attack, primarily studied within classification tasks, involves the malicious alteration of a data sample \(x\) with its label \(y\) into a new sample \(x^*\) that maximizes the loss function \(\ell(x^*; y)\) \cite{yuan2019adversarial, xu2019adversarialattacksdefensesimages,mirzaei2022scan}. This new sample \(x^*\) is known as an adversarial example of \(x\), and the difference (\(x^*\) - \(x\)) is referred to as adversarial perturbation. To prevent significant semantic changes, the \(l_p\) norm of the adversarial noise is restricted by an upper limit \(\epsilon\). Specifically, an adversarial example \(x^*\) must satisfy the condition \(x^* = \arg \max_{x^{\prime}: \|x - x^{\prime}\|_p \leq \epsilon} \ell(x^{\prime}; y)\). A widely used and effective attack method is the Projected Gradient Descent (PGD) \cite{madry2017towards} technique, which iteratively increases the loss function by moving in the direction of the gradient sign of \(\ell(x^*; y)\) with a specified step size \(\alpha\). When adapting adversarial attacks for AD, the objective shifts from maximizing the loss value to increasing \(S(x, f)\) for normal samples and decreasing it for anomalies. The attack is formulated as \(x_0^* = x\), \(x_{t+1}^* = x_t^* + y \cdot \alpha \cdot \operatorname{sign}(\nabla_x S(x_t^*; f_\theta))\), and \(x^* = x^*_k\), where \(y=+1\) for normal samples and \(y=-1\) for anomaly samples. This approach is consistently applied to other attacks in the study \cite{yuan2019adversarial, feinman2017detecting}.

\textbf{Adversarial Robustness of Anomaly Localization:}  
Adversarial attacks aim to introduce perturbations that uniformly modify pixel scores in an anomaly map \(M\), flipping scores from 0 to 1 or vice versa. In the context of AL, the attack objective transitions from merely maximizing a loss function to specifically increasing the values of \(M(x, f)\) for normal samples while decreasing them for anomalies. The attack is formalized as follows: starting with \(x_0^* = x\), the adversarial example at each step is updated as \(x_{t+1}^* = x_t^* + Y \cdot \alpha \cdot \operatorname{sign}(\nabla_x M(x_t^*; f_\theta))\), and the final adversarial example is \(x^* = x_k^*\). Here, \(Y_{i, j} = +1\) for normal pixels and \(Y_{i, j} = -1\) for anomaly pixels, where \(i, j\) denote pixel coordinates. This strategy is extended to other advanced attack methods throughout the study \cite{croce2024robust}. Despite its significance, adversarial robustness in anomaly localization remains underexplored. Enhancing robustness is particularly critical for applications requiring spatial precision, such as medical diagnostics and industrial monitoring systems.

\section{Theoretical Insights} \label{theory}

\begin{figure}[h]
  \begin{center}
    \includegraphics[width=1\linewidth]{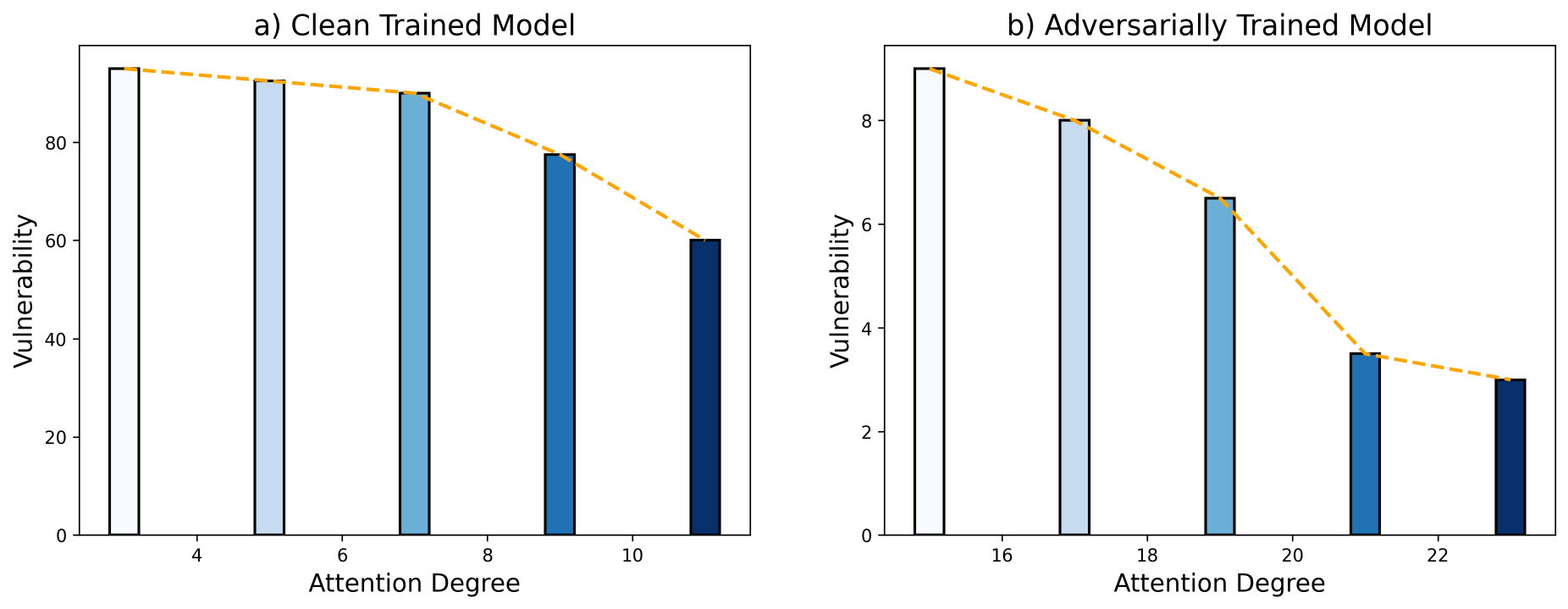}
    \caption{The figure demonstrates how increasing the average last-layer attention degree in the ViT-base architecture reduces vulnerability to adversarial attacks. Specifically, the BraTS dataset images were clustered into five groups based on their attention degree values (x-axis). The y-axis represents Vulnerability, measured as the absolute difference in the model's localization performance (AUROC\%) for each cluster before and after a PGD attack. (a) illustrates the decreasing trend in a clean-trained model. (b) shows a similar decreasing trend in the adversarially trained model, where the attention degree is relatively higher than in the clean model.}

    \label{fig:theory figure}
  \end{center}
\end{figure}

Here, we want to show why a self-attention layer, in which each token attends to a higher number of other tokens is more adversarially robust compared to the case that the attention is distributed over a smaller number of tokens. Let's consider the self-attention output embedding of the $i$th token in the $l$-th layer as $x_i^{(l)}$, and such embedding after the perturbation is applied to the input as $\hat{x}_i^{(l)}$. We are seeking to set $x_i^{(2)} - \hat{x}_i^{(2)} = r_i$, where $r_i$ is the desired {\it adversarial direction} that most influences the  layers next to the first layer of the transformer. According to the self-attention:

$$
x_i^{(l+1)} = V^{(l)}. softmax(q_i^{(l)\top }K/\sqrt{d}), 
$$
where $q_i^{(l)} = W_q x_i^{(l)}$, $v_j^{(l)}$ the $j$-th column of $V^{(l)}$ be $v_j^{(l)} = W_v x_j^{(l)}$,  $k_j^{(l)}$ the $j$-th column of K be $k_j^{(l)} = W_k x_j^{(l)}$, and $d$ be the embedding dimension. For simplicity, we assume that the dimensions of embeddings of key, query, and values are all the same and are equal to the dimension of the original embeddings. 
Now, \( W_v (X^{(1)} S - \hat{X}^{(1)} \hat{S}) = R \), where the columns of \( R \) and \( X^{(l)} \) are represented by \( r_i \) and \( x_i^{(l)} \), and the columns in \( S \) and \( \hat{S} \) correspond to the softmax values of each query token calculated for \( X \) and the perturbed inputs \( \hat{X} \), respectively. Therefore,
\begin{align*}
& W_v (X^{(1)} S - \hat{X}^{(1)} S) = \\
& W_v (X^{(1)}S - \hat{X}^{(1)} \hat{S} + \hat{X}^{(1)} \hat{S} - \hat{X}^{(1)}S) = \\
& R + W_v \hat{X}^{(1)} (\hat{S} - S) = R + W_v \hat{X}^{(1)} \Delta S.
\end{align*} 
where $\Delta S := \hat{S} - S$. Hence, we have:
\begin{align*}
X^{(1)} - \hat{X}^{(1)} = \underbrace{W_v^{+} (R + W_v \hat{X}^{(1)} \Delta S)}_{U} S^{+}.
\end{align*} 
Where \( W_v^{+} \) and \( S^+ \) are the pseudo-inverses of the matrices \( W_v \) and \( S \), respectively. Note that for the $\delta := X^{(1)} - \hat{X}^{(1)}$ perturbation in the input, in order to make the required perturbation in the output of the network, we get $\| \delta \| \leq \lambda_{max} \| U \|$, where $\lambda_{max}$ is the maximum eigenvalue of $S^+$. But we note that for a stochastic matrix $S$, the maximum eigenvalue is 1. Therefore, for the inverse matrix, $S^+$, the maximum eigenvalue would be larger than 1. As a result, we would expect the $\| \delta \|$ to grow proportional to $\lambda_{max} \geq 1$. We note that the smaller $\| \delta \|$ would imply more vulnerability as it becomes easier to attack the network with a smaller perturbation length. On the other hand, for a purely localized $S$, where each token only attends to itself $\lambda_{max} = 1$, showing that the increase in $\| \delta \|$ would likely be small compared to the general case, leading to more vulnerability. Therefore, we conclude that a localized and overly sparse attention map would typically imply an easier attack on the network and hence higher vulnerability against adversarial attacks. Sparse attention values may also resemble what happens in convolutional networks that have small receptive fields.
\section{Methodology} \label{method}

\begin{figure*}[t]
  \begin{center}
    \includegraphics[width=1\linewidth]{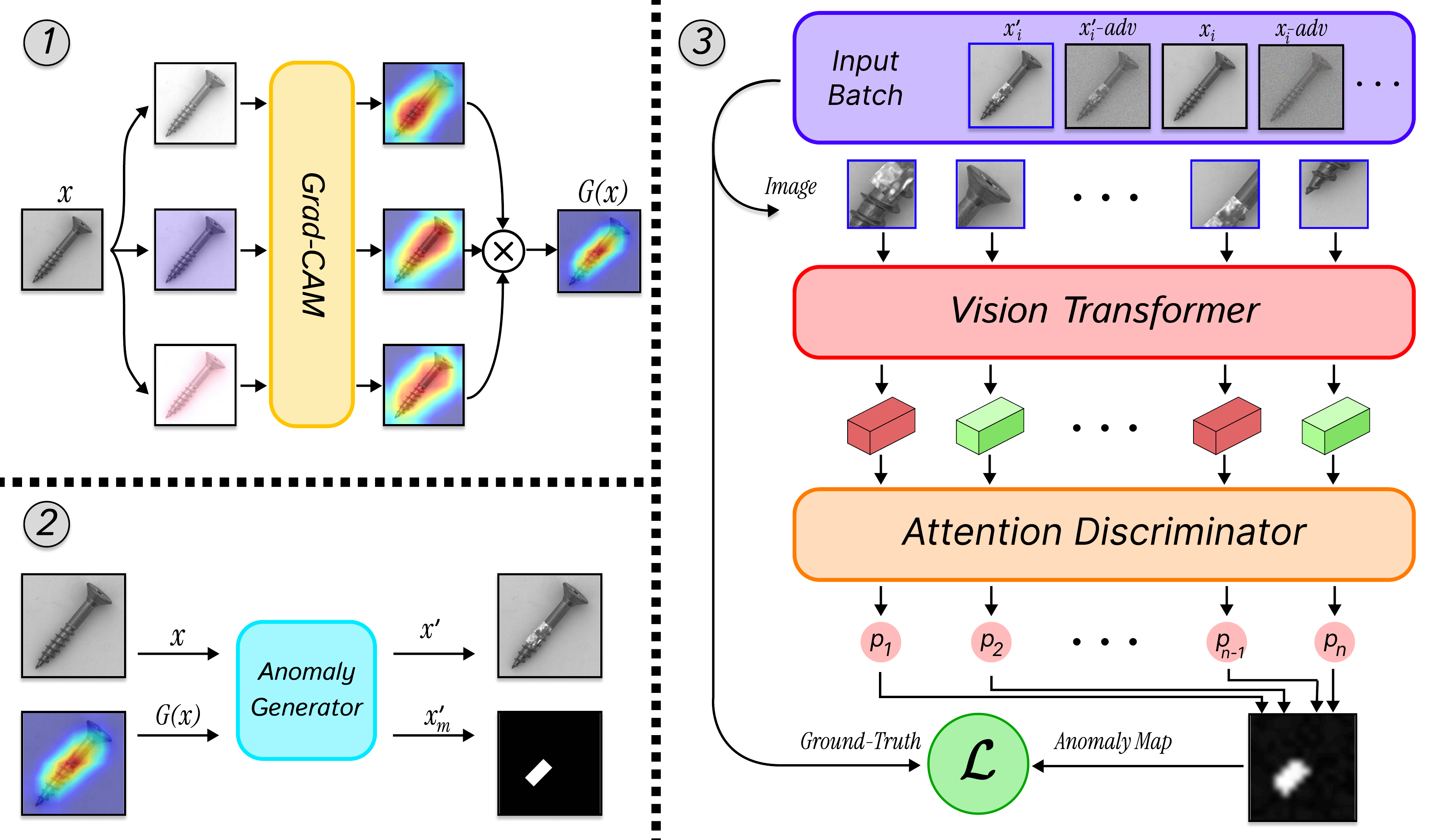}
    \caption{Overview of the PatchGuard framework. (1) We use Grad-CAM to identify regions likely to belong to the foreground in an image. By applying $k_{\text{soft}}=3$ augmentations to the image $x$, we generate a combined saliency map $G(x)$. (2) The input $x$ and its saliency map $G(x)$ are then passed to the anomaly generator to produce an anomaly sample $x'$ along with its ground-truth mask $x'_m$. (3) For each normal sample $x_i$, the input batch includes the image, its corresponding anomaly version $x'_i$, and their adversarially attacked variants, along with their ground-truth masks. Each batch sample is processed through a Vision Transformer (ViT), where the Attention Discriminator assigns an anomaly score $p_i$ to each patch embedding. The final anomaly map is constructed from these scores and trained with a novel loss $\mathcal{L}$ to replicate the ground truth.}

    \label{fig:main figure}
  \end{center}
\end{figure*}

\begin{table*}[t]
\caption{Detection performance of various AD methods, including PatchGuard, across benchmark datasets. Results are shown as ``\graytext{Clean /} Adversarial'', representing performance on ``Clean'' and ``Adversarial'' data (PGD-1000, $\epsilon = \frac{8}{255}$, measured by AUROC \% ), respectively.}

     \resizebox{ \linewidth}{!}{\begin{tabular}{@{}ccccccccccccc@{}} 

    \specialrule{1.5pt}{\aboverulesep}{\belowrulesep}
    \multirow{2}{*}{\textbf{Dataset}} & \multicolumn{10}{c}{\textbf{Method}}
    \\  \cmidrule(lr{0pt}){2-12} 
    &  \textbf{DRAEM} & \textbf{PatchCore} & \textbf{SimpleNet} &   \textbf{ReContrast} &  \textbf{GeneralAD} & \textbf{GLAD} & \textbf{GLASS} &\textbf{PrincipaLS*} & \textbf{ZARND*} & \textbf{RODEO*}  & \textbf{Ours} \\

    \specialrule{1.5pt}{\aboverulesep}{\belowrulesep}

    \noalign{\vskip 5pt} 
    
    \textbf{MVTec AD} & \graytext{98.0 /} 5.1 & \graytext{99.1 /} 3.9 & \graytext{99.6 /} 2.3 & \graytext{99.5 /} 0.0 & \graytext{99.2 /}  13.7 & \graytext{99.3 /} 11.8 & \graytext{99.9 /} 6.7 & \graytext{63.8 /} 10.6 & \graytext{71.6 /} 11.0 & \graytext{61.5 / } 4.9 & \cellcolor{blue!15}\graytext{88.1 /} 71.1 \\

    \noalign{\vskip 3pt} 
    \cmidrule(l{0pt}r){1-1} \cmidrule(lr{0pt}){2-12}
    \noalign{\vskip 3pt} 

     \textbf{VisA} & \graytext{88.7 /} 3.8 & \graytext{95.1 /} 1.7 & \graytext{96.2 /} 0.2 & \graytext{97.5 /} 6.7 & \graytext{95.9 /} 14.5 & \graytext{99.5 /}  8.6 & \graytext{98.8 /} 7.2 & \graytext{58.9 /} 9.3 & \graytext{67.7 /} 9.3 & \graytext{63.1 / } 3.8 & \cellcolor{blue!15}\graytext{88.5 /} 74.3 \\

    \noalign{\vskip 3pt} 
    \cmidrule(l{0pt}r){1-1} \cmidrule(lr{0pt}){2-12}
    \noalign{\vskip 3pt} 
    
    \textbf{BTAD}& \graytext{87.1 /} 4.9 & \graytext{96.9 /} 2.4 & \graytext{95.1 /} 4.6 & \graytext{95.7 /} 2.1 & \graytext{97.0 /} 10.9 & \graytext{98.5 /}  7.6 & \graytext{97.4 /} 8.3 & \graytext{60.1 /} 7.8 & \graytext{69.7 /} 8.3 & \graytext{60.0 / } 5.8 & \cellcolor{blue!15}\graytext{85.3 /} 82.1 \\

    \noalign{\vskip 3pt} 
    \cmidrule(l{0pt}r){1-1} \cmidrule(lr{0pt}){2-12}
    \noalign{\vskip 3pt} 

     \textbf{MPDD} & \graytext{94.3 /} 2.8 & \graytext{91.3 /} 6.8 & \graytext{96.6 /} 7.0 & \graytext{94.5 /} 4.5 & \graytext{98.0 /} 12.8 & \graytext{97.5 /} 9.7 & \graytext{99.6 /} 9.1 & \graytext{71.9 /} 15.7 & \graytext{65.7 /} 12.2& \graytext{68.8 / } 7.4 & \cellcolor{blue!15}\graytext{85.4 /} 68.6 \\

    \noalign{\vskip 3pt} 
    \cmidrule(l{0pt}r){1-1} \cmidrule(lr{0pt}){2-12}
    \noalign{\vskip 3pt}

    \textbf{WFDD} & \graytext{94.0 /} 7.1 & \graytext{96.3 /} 3.9 & \graytext{98.8 /} 5.1 & \graytext{97.6 /} 3.3 & \graytext{99.0 /} 10.7 & \graytext{99.1 /} 12.5 & \graytext{100 /} 6.0 & \graytext{66.3 /} 14.1 & \graytext{76.1 /} 16.3 & \graytext{71.2 / } 7.8 & \cellcolor{blue!15}\graytext{84.2 /} 65.0 \\

    \noalign{\vskip 3pt} 
    \cmidrule(l{0pt}r){1-1} \cmidrule(lr{0pt}){2-12}
    \noalign{\vskip 3pt}

    \textbf{DTD-Synthetic} & \graytext{97.1 /} 3.6 & \graytext{65.8 /} 4.0 & \graytext{96.8 /} 2.4 & \graytext{98.2 /} 9.1 & \graytext{97.9 /} 16.3 & \graytext{98.6 /} 10.3  & \graytext{99.1 /} 6.1 & \graytext{77.9 /} 9.3 & \graytext{70.3 /} 13.7& \graytext{72.8 / } 12.0 & \cellcolor{blue!15}\graytext{91.5 /} 60.5 \\

    \noalign{\vskip 3pt} 
    \cmidrule(l{0pt}r){1-1} \cmidrule(lr{0pt}){2-12}
    \noalign{\vskip 3pt}

    \textbf{Head-CT} & \graytext{72.1 /} 8.1 & \graytext{92.4 /} 5.1 & \graytext{84.9 /} 4.8 & \graytext{86.1 /} 6.3 & \graytext{89.1 /} 9.1 & \graytext{93.1 /} 14.2 & \graytext{94.7 /} 8.7 & \graytext{73.2 /} 16.0 & \graytext{64.3 /} 9.7 & \graytext{85.7 / } 61.0 & \cellcolor{blue!15}\graytext{97.3 /} 90.4 \\

    \noalign{\vskip 3pt} 
    \cmidrule(l{0pt}r){1-1} \cmidrule(lr{0pt}){2-12}
    \noalign{\vskip 3pt}

     \textbf{BraTS2021} & \graytext{62.3 /} 1.4 & \graytext{91.6 /} 6.3 & \graytext{82.5 /} 6.1 & \graytext{82.4 /} 3.1 & \graytext{88.3 /} 13.0 & \graytext{92.3 /} 12.0 & \graytext{95.7 /} 7.2 & \graytext{75.3 /} 8.7 & \graytext{66.7 /} 14.0& \graytext{89.1 / } 64.7 & \cellcolor{blue!15}\graytext{94.3 /} 81.0 \\

    \noalign{\vskip 3pt}
    \specialrule{1.5pt}{\aboverulesep}{\belowrulesep}
    \noalign{\vskip 3pt} 

    \textit{\textbf{ Average}} & \graytext{86.7 /} 4.6 & \graytext{91.0 /} 4.2 & \graytext{93.8 /} 4.0 & \graytext{93.9 /} 4.3 & \graytext{95.5 /} 12.6 & \graytext{97.2 /} 10.8 & \graytext{98.1 /} 7.4 & \graytext{68.4 /} 11.4 & \graytext{69.0 /} 11.8 & \graytext{71.5 / } 20.9 & \cellcolor{blue!15}\graytext{89.3 /} 74.1 \\

    \noalign{\vskip 3pt}

    \specialrule{1.5pt}{\aboverulesep}{\belowrulesep}

     \end{tabular}}
    \label{tab:Main_table}
    \footnotesize{\scriptsize $^*$These works incorporated adversarial training into their proposed AD methods.}

\end{table*}

\begin{table*}[t]
    \caption{Localization performance of various AL methods, including PatchGuard, across benchmark datasets. Results are shown as ``\graytext{Clean /} Adversarial'', representing performance on ``Clean'' and ``Adversarial'' data (PGD-1000, $\epsilon = \frac{8}{255}$, measured by AUROC \% ), respectively.}
     
     \resizebox{ \linewidth}{!}{\begin{tabular}{@{}ccccccccccccc@{}} 

    \specialrule{1.5pt}{\aboverulesep}{\belowrulesep}
    \multirow{2}{*}{\textbf{Dataset}} & \multicolumn{10}{c}{\textbf{Method}}
    \\  \cmidrule(lr{0pt}){2-12} 
    &  \textbf{DRAEM} & \textbf{PatchCore} & \textbf{SimpleNet} &   \textbf{ReContrast} &  \textbf{GeneralAD} & \textbf{GLAD} & \textbf{GLASS} & \textbf{MuSc} & \textbf{ReConPatch} & \textbf{RealNet}  & \textbf{Ours} \\

    \specialrule{1.5pt}{\aboverulesep}{\belowrulesep}

    \noalign{\vskip 5pt} 
    
    \textbf{MVTec AD} & \graytext{97.3 /} 3.9 & \graytext{98.1 /} 4.2 & \graytext{98.1 /} 6.8 & \graytext{98.4 /} 4.8 & \graytext{97.7 /} 12.3 & \graytext{98.7 /}  15.3 & \graytext{99.3 /} 8.1 & \graytext{97.3 /} 4.6 & \graytext{99.2 /} 2.8 & \graytext{99.0 / } 8.7 & \cellcolor{blue!15}\graytext{92.7 /} 73.8 \\

    \noalign{\vskip 3pt}
    \cmidrule(l{0pt}r){1-1} \cmidrule(lr{0pt}){2-12}
    \noalign{\vskip 3pt}

     \textbf{VisA} & \graytext{90.7 /} 5.6 & \graytext{98.5 /} 3.0 & \graytext{97.4 /} 4.8 & \graytext{98.2 /} 7.1 & \graytext{99.0 /} 7.6 & \graytext{98.6 /}  13.4 & \graytext{98.8 /} 7.0 & \graytext{98.8 /} 3.8 & \graytext{99.2 /} 1.7 & \graytext{98.8 / } 6.2 & \cellcolor{blue!15}\graytext{96.9 /} 85.2 \\

    \noalign{\vskip 3pt} 
    \cmidrule(l{0pt}r){1-1} \cmidrule(lr{0pt}){2-12}
    \noalign{\vskip 3pt} 
    
    \textbf{BTAD}& \graytext{89.0 /} 7.1 & \graytext{92.6 /} 6.3 & \graytext{95.2 /} 3.7 & \graytext{96.8 /} 6.4 & \graytext{96.7 /} 9.3 & \graytext{98.2 /}  14.5 & \graytext{97.0 /} 5.7 & \graytext{97.3 /} 5.3 & \graytext{97.5 /} 3.0 & \graytext{97.9 / } 9.2 & \cellcolor{blue!15}\graytext{93.2 /} 73.0 \\

    \noalign{\vskip 3pt} 
    \cmidrule(l{0pt}r){1-1} \cmidrule(lr{0pt}){2-12}
    \noalign{\vskip 3pt}

     \textbf{MPDD} & \graytext{90.7 /} 4.7 & \graytext{98.5 /} 1.9 & \graytext{97.4 /} 7.3 & \graytext{94.6 /} 3.6 & \graytext{98.1 /} 14.9 & \graytext{98.7 /} 10.8 & \graytext{99.4 /} 6.3 & \graytext{95.5 /} 6.0 & \graytext{96.7 /} 2.6 & \graytext{98.2 /} 7.5 & \cellcolor{blue!15}\graytext{93.8 /} 86.6 \\

    \noalign{\vskip 3pt} 
    \cmidrule(l{0pt}r){1-1} \cmidrule(lr{0pt}){2-12}
    \noalign{\vskip 3pt}

    \textbf{WFDD} & \graytext{95.8 /} 3.9 & \graytext{98.1 /} 2.1 & \graytext{98.0 /} 3.4 & \graytext{97.3 /} 5.1 & \graytext{98.7 /} 13.5 & \graytext{97.6 /} 11.3 & \graytext{98.9 /} 7.6 & \graytext{97.0 /} 4.1 & \graytext{98.3 /} 4.9 & \graytext{98.4 / } 6.8 & \cellcolor{blue!15}\graytext{94.6 /} 71.6 \\

    \noalign{\vskip 3pt} 
    \cmidrule(l{0pt}r){1-1} \cmidrule(lr{0pt}){2-12}
    \noalign{\vskip 3pt}

    \textbf{DTD-Synthetic} & \graytext{96.1 /} 4.8 & \graytext{95.6 /} 6.7 & \graytext{97.0 /} 2.9 & \graytext{96.8 /} 7.6 & \graytext{97.9 /} 10.8 & \graytext{97.3 /} 13.8 & \graytext{98.1 /} 7.0 & \graytext{97.8 /} 5.7 & \graytext{96.3 /} 3.9 & \graytext{98.0 / } 8.3 & \cellcolor{blue!15}\graytext{95.9 /} 74.8 \\

    \noalign{\vskip 3pt} 
    \cmidrule(l{0pt}r){1-1} \cmidrule(lr{0pt}){2-12}
    \noalign{\vskip 3pt}

    \textbf{Head-CT} & \graytext{86.4 /} 8.9 & \graytext{87.3 /} 6.1 & \graytext{89.7 /} 0.8 & \graytext{90.1 /} 8.6 & \graytext{91.7 /} 18.3 & \graytext{93.4 /} 16.7 & \graytext{92.2 /} 9.1 & \graytext{85.2 /} 8.5 & \graytext{89.7 /} 6.6 & \graytext{94.8 / } 7.3 & \cellcolor{blue!15}\graytext{93.6 /} 89.3 \\

    \noalign{\vskip 3pt} 
    \cmidrule(l{0pt}r){1-1} \cmidrule(lr{0pt}){2-12}
    \noalign{\vskip 3pt}

     \textbf{BraTS2021} & \graytext{82.2 /} 6.1 & \graytext{96.9 /} 4.5 & \graytext{94.7 /} 6.7 & \graytext{95.3 /} 11.1 & \graytext{96.8 /} 14.7 & \graytext{95.1 /} 13.5 & \graytext{94.0 /} 6.3 & \graytext{67.5 /} 6.3 & \graytext{97.5 /} 5.9 & \graytext{96.9/} 10.2 & \cellcolor{blue!15}\graytext{97.7 /} 94.5 \\

    \noalign{\vskip 3pt}
    \specialrule{1.5pt}{\aboverulesep}{\belowrulesep}
    \noalign{\vskip 3pt} 

    \textit{\textbf{ Average}} & \graytext{91.0 /} 5.6 & \graytext{95.7 /} 4.3 & \graytext{95.9 /} 4.5 & \graytext{95.9 /} 6.7 & \graytext{97.0 /} 12.6 & \graytext{97.2 /} 13.6 & \graytext{97.2 /} 7.1 & \graytext{92.0 /} 5.5 & \graytext{96.8 /} 3.9 & \graytext{97.7 / } 8.0 & \cellcolor{blue!15}\graytext{94.8 /} 81.1 \\

    \noalign{\vskip 3pt}

    \specialrule{1.5pt}{\aboverulesep}{\belowrulesep}

     \end{tabular}}
    \label{tab:Main_table_localization}

\end{table*}

\textbf{Overview.}
Current AD and AL methods perform poorly when subjected to adversarial attacks. Previous studies have demonstrated the effectiveness of OE in enhancing robustness against such attacks \cite{azizmalayeri2022your, chen2020robust, chen2021atom, mirzaeirodeo}. Additionally, ViTs have shown promising results in standard (clean) AD and AL tasks \cite{cohen2021transformaly, strater2024generalad, li2024musc, jeong2023winclip}, and our findings suggest their potential for adversarial settings with targeted modifications of attention degrees. Building on these insights and our theoretical foundations, we propose PatchGuard, a novel method designed to enhance adversarial robustness in AD and AL. The following subsections detail the key components of PatchGuard.

\subsection{Foreground-Aware Pseudo-Anomaly Generation} \label{foreground}

To generate anomalous data from normal samples, it is essential to apply targeted distortions that ensure the image becomes anomalous \cite{caron2020unsupervised, chen2020simple, chen2021exploring, grill2020bootstrap}. Additionally, the generated pseudo-anomalies must satisfy two key requirements: (1) inclusion of localization maps to enhance the AL task, and (2) close alignment with the normal sample distribution \cite{mirzaeirodeo, mirzaei2024universal}. Meeting these requirements simultaneously necessitates precise but minimal alterations within the foreground. Thus, the proposed method must reliably identify a region of the image that we can confidently attribute to the foreground.

To achieve this, we leverage insights from \cite{noohdani2024decompose}, which indicate that pretrained models benefit from style-agnostic processing to improve classification accuracy. To this end, we first introduce \textit{soft transformations}, which are commonly used in self-supervised learning studies due to their ability to preserve semantic content \cite{chen2020simple,he2020momentum,grill2020bootstrap,caron2020unsupervised,chen2021exploring}. Next, we apply Grad-CAM \cite{gradcam} to generate a saliency map using a standard pre-trained ResNet18 \cite{resnet} model on a given input. Specifically, for a normal sample \( x \), we randomly select \( k_{\text{soft}} \) soft transformations \( t_1^s, \dots, t_{k_{\text{soft}}}^s \). We then compute saliency maps for \( x, t_1^s(x), \dots, t_{k_{\text{soft}}}^s(x) \) and take their element-wise product to produce a style-invariant saliency map, \( G(x) \). As shown in Figure \ref{fig:main figure} (part \raisebox{.5pt}{\textcircled{\raisebox{-.9pt} {1}}}), \( G(x) \) effectively assigns higher values to the foreground region.

To distort the obtained region, we first randomly sample an anchor point from it. We then sample \( k_{\text{hard}} \) hard transformations \( t_1^h, \dots, t_{k_{\text{hard}}}^h \), which are typically associated with altering semantic integrity, as highlighted in prior research \cite{zhang2018mixup,Akbiyik2019DataAI,ghiasi2020simple,tack2020csi,sohn2020learning,kalantidis2020hard,li2021cutpaste,sinha2021negative,DBLP:conf/ijcai/ChenXLQZTZM21,devries2017cutout,yun2019cutmix,mirzaei2025contrastiveteacherstudentframeworknovelty}. We then design a rectangular mask centered on this chosen point, with its width \( w \) and height \( h \) satisfying \( w \sim U(0.05W, 0.3W) \) and \( h \sim U(0.05H, 0.3H) \), where \( W \) and \( H \) are the image’s width and height. Next, we rotate the mask by an angle of \( \theta \sim U(-45^\circ, 45^\circ) \). Finally, we sequentially apply the hard transformations only to the region within this mask, leaving the remaining unmasked areas of the image unchanged. Mathematically, the final anomaly is obtained by applying
\begin{equation}
    x' = t_1^h( \dots( t_{k_{\text{hard}}}^h(x_{\text{masked}})\dots) + (x - x_{\text{masked}}).    
\end{equation}
An additional advantage of this pseudo-anomaly crafting approach is that the location of the distorted region is precisely known, allowing us to generate an accurate localization map \( x'_m \) for AL training, where each pixel's value equals 1 if it's anomalous, and 0 otherwise. By creating an anomalous counterpart $x'$ and its corresponding localization map $x'_m$ for each normal sample $x$ in the training set, we provide the model with both the anomaly and its location, as shown in Figure \ref{fig:main figure} (part \raisebox{.5pt}{\textcircled{\raisebox{-.9pt} {2}}}). This pairing equips the model with the information needed to improve its localization accuracy, as we will discuss in Section \ref{vit}. For details on the soft and hard transforms, and ablations on the hypers used in this approach, refer to Table \ref{tab:pseudo anomaly table} and Appendix \ref{appendix:augmentation}.
\subsection{Anomaly Detection and Localization with Vision Transformer} \label{vit}
To enhance adversarial robustness through the effective use of pseudo-anomaly samples and their corresponding localization maps, we employ a Vision Transformer (ViT) model. This model generates output embeddings that are processed by an attention-based discriminator network, which uses a multi-head attention (MLH) mechanism followed by a shared-weight multi-layer perceptron (MLP). The discriminator independently assigns an anomaly score to each patch in the input, providing a patch-level anomaly prediction as shown in Figure \ref{fig:main figure} (part \raisebox{.5pt}{\textcircled{\raisebox{-.9pt} {3}}}).

The model is trained using a combination of patch-level cross-entropy (CE) and an additional regularization term \(R\), designed to encourage a higher degree of attention in a specified ViT layer. As illustrated in Figure \ref{fig:theory figure} and the theory in Section \ref{theory}, this regularizer is expected to further enhance adversarial robustness. The cross-entropy loss component aligns the predicted anomaly scores with ground truth labels from the localization map, $x_m$, where a patch label equals 1 if more than \(5\%\) of the pixels in the corresponding patch in \(x_m\) are anomalous, and 0 otherwise:
\begin{align}
    M &= \text{MLP}(\text{MLH}(x)) \\
    \mathcal{L}_{\text{CE}}(x, x_m) &= \sum_{i, j \in x_{\text{patches}}} \text{CE}(M_{(i, j)}, x_{m, (i, j)})
\end{align}
Here, \(M\) denotes the predicted AL map, consisting of anomaly scores for each patch, \(x\) is any input sample (normal or anomaly), and \(i, j\) are the indices for individual patches. The cross-entropy loss term \(\mathcal{L}_{\text{CE}}\) ensures alignment between the patch \(M_{(i,j)}\) and the ground truth anomaly map \(x_{m,(i,j)}\), promoting precise anomaly localization. The final AL map is obtained by upsampling \(M\) to the original image size, and we use the average of the top-\(k\) patch-level anomaly scores as the image-level anomaly score.

To further improve the robustness of the model, we introduce a regularization term \(R\), which encourages an increased attention degree in a specified layer \(\ell\) of the ViT, chosen here as the final layer. This regularization term is formulated as:
\begin{equation}
R(x) = \frac{1}{\sum_{i} \sum_{j} \sum_{k} A_{ijk} \cdot \mathbbm{1}(A_{ijk} \leq \delta)}
\end{equation} 
Here, \( A_{ijk} \) represents the attention coefficient corresponding to the \(i\)-th head, the \(j\)-th output patch, and the \(k\)-th input patch. Regularization will increase the values of attention coefficients that are below the threshold \( \delta \). For a specific output patch, since a softmax function is applied to its corresponding attention coefficients, increasing the lower values naturally decreases the higher ones. With \( \delta=\frac{1}{\text{(number of input patches)}} \), the coefficient distribution gradually shifts toward a uniform distribution at this threshold, representing the maximum attention degree a patch can achieve.
The total loss function \(\mathcal{L}\) used for training the model is the sum of the cross-entropy and the regularization term:
\begin{equation}
    \mathcal{L}(x,x_m) = \mathcal{L}_{\text{CE}}(x,x_m) + \alpha R(x)
\end{equation}
where \(\alpha\) is a scaling factor that controls the strength of the regularization. This combined objective encourages the model to not only accurately predict anomaly locations but also maintain an increased level of attention, enhancing both detection performance and adversarial robustness. It is then used in adversarial training, as explained in Section \ref{adversarial training}. Comprehensive ablation studies of hyperparameters including $\delta$ and $\alpha$ are provided in Appendix \ref{appendix:vit ablation} and Section \ref{sec:ablation}.

\subsection{Adversarial Training and Testing} \label{adversarial training}

Given an input sample \( x \), an adversarial example \( x_{adv} \) is generated by introducing a perturbation \( \delta^* \), optimized to maximize the final loss: 
\begin{align}
\delta^* = \underset{\|\delta\|_{\infty} \leq \epsilon}{\mathrm{argmax}}\, \mathcal{L}(x + \delta, x_m), \quad x_{adv} = x + \delta^* 
\end{align}
These adversarial examples are then used in the training process alongside the original examples. The adversarial training objective is framed as a min-max problem, optimizing the model parameters \( \theta \) to minimize the expected loss over both clean and adversarial examples:
\begin{align}
\min_{\theta} \mathbb{E}_{(x, y) \in \mathcal{B} } \left[ \max_{\|\delta\|_\infty \leq \epsilon} \mathcal{L}(x + \delta, x_m; \theta) \right].
\end{align}

\section{Experiments} \label{sec:experiments}

To demonstrate the effectiveness of PatchGuard, we conduct extensive experiments assessing its performance alongside several established anomaly detection methods on multiple benchmarks for anomaly detection and localization. Our evaluation includes both adversarially trained models and those trained under standard conditions. To ensure a fair comparison, we assess all methods under both ``Clean" and ``Adversarial" test conditions, as shown in Table \ref{tab:Main_table} and \ref{tab:Main_table_localization}. In the Clean setting, test samples are unaltered, whereas the Adversarial setting includes samples perturbed by the PGD attack \cite{madry2017towards}. Additional results on PatchGuard's performance against alternative attacks are provided in Section \ref{ablation attack}. Datasets are detailed in Appendix \ref{appendix:dataset details}.

\textbf{Evaluation Details.} In the adversarial setting, we evaluate each method using the $\ell_\infty$ PGD-1000 attack with $\epsilon = \frac{8}{255}$. Detection and localization results are shown in Tables \ref{tab:Main_table} and \ref{tab:Main_table_localization}, respectively. $\ell_2$ norm evaluations are in Appendix \ref{appendix: various attacks diverse epsilon}.


\textbf{Adapting State-of-the-Art AD and AL Methods to Adversarial Settings.}
Prior to developing our novel PatchGuard framework, we first investigated whether existing state-of-the-art methods in AD and AL could be enhanced through adversarial training—a technique where models are trained on adversarial examples to improve their robustness. In Appendix \ref{appendix:detail-adaptation-sota-method-to-adv-training}, we provide a detailed methodology for generating adversarial examples for each method and describe our optimized approach for improving their resilience. The empirical results, presented in Table \ref{tab:adapting clean methods table}, reveal that even after incorporating adversarial training, current SOTA AD and AL methods remain vulnerable to adversarial attacks and exhibit suboptimal performance. These limitations motivated the development of PatchGuard, our proposed solution.
\begin{table}[t]
    \caption{Performance of prior methods adapted for adversarial settings, shown as ``\graytext{Clean /} Adversarial'' (PGD-1000, $\epsilon = \frac{8}{255}$).}
    
    \resizebox{ \columnwidth}{!}
    {\begin{tabular}{@{}cccccc} 
    
    \specialrule{1.5pt}{\aboverulesep}{\belowrulesep}
    \multirow{2}{*}{\textbf{Method}} & \multirow{2}{*}{\textbf{Task}} & \multicolumn{4}{c}{\textbf{Dataset}} \\
      \cmidrule(lr){3-6}

&& \textbf{MVTec AD}& \textbf{VisA}& \textbf{BTAD} & \textbf{BraTS2021}\\ 

    \specialrule{1.5pt}{\aboverulesep}{\belowrulesep} 

    \noalign{\vskip 1.5pt}

   \multirow{2}{*}{PatchCore} & AD & \textcolor{gray}{92.3 / }14.3 & \textcolor{gray}{88.4 / }10.8 & \textcolor{gray}{87.1 / }12.7 & \textcolor{gray}{88.6 / }16.1\\

   \cmidrule(lr){3-6}

   & AL & \textcolor{gray}{89.1 / }10.3 & \textcolor{gray}{87.6 / }8.8 & \textcolor{gray}{85.0 / }11.5 & \textcolor{gray}{86.7 / }13.6\\

   \noalign{\vskip 1.5pt} 
    \specialrule{1pt}{\aboverulesep}{\belowrulesep}
    \noalign{\vskip 1.5pt}

   \multirow{2}{*}{ReContrast} & AD & \textcolor{gray}{93.8 / }5.1 & \textcolor{gray}{93.7 / }16.7 & \textcolor{gray}{89.9 / }13.4 & \textcolor{gray}{78.2 / }13.1\\

   \cmidrule(lr){3-6}

   & AL & \textcolor{gray}{94.5 / }17.2 & \textcolor{gray}{92.5 / }18.2 & \textcolor{gray}{88.6 / }17.9 & \textcolor{gray}{90.8 / }14.2\\

   \noalign{\vskip 1.5pt} 
    \specialrule{1pt}{\aboverulesep}{\belowrulesep}
    \noalign{\vskip 1.5pt} 

   \multirow{2}{*}{SimpleNet} & AD & \textcolor{gray}{91.1 / }18.1 & \textcolor{gray}{89.2 / }14.9 & \textcolor{gray}{88.6 / }12.7 & \textcolor{gray}{75.3 / }14.2\\

   \cmidrule(lr){3-6}

   & AL & \textcolor{gray}{94.1 / }19.5 & \textcolor{gray}{89.4 / }15.8 & \textcolor{gray}{88.2 / }11.7 & \textcolor{gray}{90.1 / }17.0\\

   \noalign{\vskip 1.5pt} 
    \specialrule{1.5pt}{\aboverulesep}{\belowrulesep}
    \noalign{\vskip 1.5pt} 
    
    \multirow{2}{*}{GeneralAD} & AD & \textcolor{gray}{84.6 / }24.7 & \textcolor{gray}{81.2 / }22.1 & \textcolor{gray}{82.9 / }20.6 & \textcolor{gray}{80.7 / }19.9\\

   \cmidrule(lr){3-6}

   & AL & \textcolor{gray}{83.7 / }21.8 & \textcolor{gray}{83.1 / }17.3 & \textcolor{gray}{84.7 / }18.3 & \textcolor{gray}{79.6 / }21.0\\

   \noalign{\vskip 1.5pt} 
    \specialrule{1.5pt}{\aboverulesep}{\belowrulesep}

    \multirow{2}{*}{DRAEM} & AD & \textcolor{gray}{83.0 / }19.2 & \textcolor{gray}{81.7 / }16.8 & \textcolor{gray}{79.1 / }17.0 & \textcolor{gray}{57.4 / }12.2\\

   \cmidrule(lr){3-6}

   & AL & \textcolor{gray}{86.1 / }17.5 & \textcolor{gray}{78.6 / }15.0 & \textcolor{gray}{80.9 / }18.6 & \textcolor{gray}{73.6 / }14.9\\

    \noalign{\vskip 1.5pt} 
    \specialrule{1.5pt}{\aboverulesep}{\belowrulesep}

    \multirow{2}{*}{GLASS} & AD & \textcolor{gray}{84.2 / }26.7 & \textcolor{gray}{83.5 / }19.9 & \textcolor{gray}{84.7 / }25.4 & \textcolor{gray}{83.7 / }20.1\\

   \cmidrule(lr){3-6}

   & AL & \textcolor{gray}{85.9 / }21.2 & \textcolor{gray}{87.1 / }21.7 & \textcolor{gray}{83.0 / }19.0 & \textcolor{gray}{85.8 / }23.6\\

   \noalign{\vskip 1.5pt} 
    \specialrule{1pt}{\aboverulesep}{\belowrulesep}
    \noalign{\vskip 1.5pt} 

    \multirow{2}{*}{\textit{Ours}} & AD & \textcolor{gray}{88.1 / }71.1 & \textcolor{gray}{88.5 / }74.3 & \textcolor{gray}{85.3 / }82.1 & \textcolor{gray}{94.3 / }81.0\\

   \cmidrule(lr){3-6}

   & AL & \textcolor{gray}{92.7 / }73.8 & \textcolor{gray}{96.9 /}85.2 & \textcolor{gray}{93.2  /}73.0 & \textcolor{gray}{97.7 /}94.5\\

   \noalign{\vskip 1.5pt} 
    \specialrule{1.5pt}{\aboverulesep}{\belowrulesep}
     \end{tabular}}
    \label{tab:adapting clean methods table}

\end{table}

\textbf{Analyzing Results.} As shown in Tables \ref{tab:Main_table} and \ref{tab:Main_table_localization}, prior SOTA methods such as Recontrast demonstrate substantial drops in performance under adversarial settings, despite excelling in clean conditions. Furthermore, as indicated in Table \ref{tab:adapting clean methods table}, even when adversarially trained, these methods struggle to achieve robust performance. By contrast, approaches specifically designed to address adversarial robustness, such as RODEO, still fall short when compared to PatchGuard. On average, PatchGuard improves robust detection across diverse datasets by up to 53.2\% percent, with a notable robust improvement of 68.5\% percent on the challenging MVTec dataset. Notably, we achieve a $71.8\%$ improvement in AL on the challenging VisA dataset, with details in Appendix \ref{appendix:per-class results}. As prior research indicates, a slight decrease in clean performance is generally viewed as an acceptable tradeoff for the critical gains in robustness.

\textbf{Implementation Details. } We leverage a from-scratch ViT model in our pipeline to showcase our setup's independence from pre-train datasets. However, we conduct an ablation study on this manner in Appendix \ref{appendix:pretrain ablation}. Further, we leverage PGD-10 with $\epsilon=\frac{8}{255}$ for adversarial training, but the test results are generated with the PGD-1000 attack. For more details on the implementation parameters, refer to Appendix \ref{appendix:implementation}.

\section{Ablation Studies} \label{sec:ablation}

\begin{table}[t]
\caption{Evaluation of PatchGuard's performance against advanced classification and segmentation attacks, showcasing its robustness in diverse attack scenarios.}
     \resizebox{ \linewidth}{!}{\begin{tabular}{@{}cccccccc@{}} 
    \specialrule{1.5pt}{\aboverulesep}{\belowrulesep}
    \multirow{2}{*}{\textbf{Dataset}} & \multirow{2}{*}{\textbf{Task}} & \multicolumn{4}{c}{\textbf{Detection Attacks}} & \multicolumn{2}{c}{\textbf{Localization Attacks}}
    \\  \cmidrule(lr){3-6}  \cmidrule(lr{0pt}){7-8}
    &&  \textbf{PGD-1000} & \textbf{CAA} &  \textbf{AutoAttack} & \textbf{$\text{A}^3$} & \textbf{SegPGD} & \textbf{SEA} \\

    \specialrule{1.5pt}{\aboverulesep}{\belowrulesep}

    \noalign{\vskip 5pt} 
    
    \multirow{2}{*}{\textbf{MVTec AD}} & AD & 71.1 & 74.8 & 70.8 & 69.8 & 71.9 & 69.2 \\

    \cmidrule(lr{0pt}){3-8}

    & AL & 73.8 & 76.7 & 72.9 & 70.0 & 74.6 & 71.3 \\

    \noalign{\vskip 1.5pt} 
    \specialrule{1pt}{\aboverulesep}{\belowrulesep}
    \noalign{\vskip 1.5pt} 

    \multirow{2}{*}{\textbf{VisA}} & AD & 74.3 & 78.5 & 74.6 & 73.1 & 75.0 & 73.6 \\

    \cmidrule(lr{0pt}){3-8}

    & AL & 85.2 &  87.4 & 85.5 & 84.7 & 85.0 & 83.7 \\

    \noalign{\vskip 1.5pt} 
    \specialrule{1pt}{\aboverulesep}{\belowrulesep}
    \noalign{\vskip 1.5pt} 

    \multirow{2}{*}{\textbf{BTAD}} & AD &  82.1 & 83.4 & 82.0 & 80.7 & 81.9 & 80.8 \\

    \cmidrule(lr{0pt}){3-8}

    & AL &  73.0 & 74.7 & 72.3 & 71.5 & 72.9 & 71.1 \\

    \noalign{\vskip 1.5pt} 
    \specialrule{1pt}{\aboverulesep}{\belowrulesep}
    \noalign{\vskip 1.5pt} 

    \multirow{2}{*}{\textbf{BraTS2021}} & AD & 81.0 & 84.7 & 81.9 & 80.3 & 82.0 & 81.1 \\

    \cmidrule(lr{0pt}){3-8}

    & AL & 94.5 & 95.0 & 94.2 & 93.8 & 94.8 & 93.4 \\

    \noalign{\vskip 3pt}

    \specialrule{1.5pt}{\aboverulesep}{\belowrulesep}

     \end{tabular}}
    \label{tab:advanced attack table}

\end{table}

In this section, we provided further analysis of our choices of parameters and other components of our method.

\subsection{Regularization} \label{ablation regularization}
To experimentally demonstrate the effectiveness of our regularization term on the model's robustness and the impact of the scaling factor $\alpha$ (set to 1 by default in our method), we conduct an ablation study on $\alpha$, as shown in Table \ref{tab:ablation scaling factor}. When $\alpha=0$, the model's robustness against adversarial attacks decreases due to the lack of regularization. On the other hand, higher values of $\alpha$, such as $\alpha=5$, cause the regularization term to dominate, leading to an unstable loss function and inappropriate results.
\begin{table}[t]
    \caption{Ablation study on the scaling factor $\alpha$, illustrating the effectiveness of our regularization.}
     \resizebox{ \columnwidth}{!}{\begin{tabular}{@{}cccccc@{}} 
    \specialrule{1.5pt}{\aboverulesep}{\belowrulesep}
    \multirow{2}{*}{\textbf{Scaling Factor ($\alpha$)}} & \multirow{2}{*}{\textbf{Task}} & \multicolumn{4}{c}{\textbf{Dataset}}
    \\  \cmidrule(lr{0pt}){3-6} 

    &&  \textbf{MVTec-AD} & \textbf{VisA} & \textbf{BTAD} & \textbf{BraTS2021}  \\

    \specialrule{1.5pt}{\aboverulesep}{\belowrulesep}

    \noalign{\vskip 3pt} 
    
    \multirow{2}{*}{\textbf{0}} & AD & \graytext{89.3 /} 60.7 & \graytext{90.1 /} 62.9 & \graytext{88.6 /} 74.3 & \graytext{95.2 /} 72.0 \\

    \cmidrule(lr{0pt}){3-6}

    & AL & \graytext{93.5 /} 65.3 & \graytext{95.8 /} 72.1 & \graytext{94.0 /} 59.7 & \graytext{98.2 /} 83.4 \\

    \noalign{\vskip 1.5pt} 
    \specialrule{1pt}{\aboverulesep}{\belowrulesep}
    \noalign{\vskip 1.5pt} 

    \multirow{2}{*}{\textbf{1 (Ours)}} & AD & \textcolor{gray}{88.1 / }71.1 & \textcolor{gray}{88.5 / }74.3 & \textcolor{gray}{85.3 / }82.1 & \textcolor{gray}{94.3 / }81.0\\

   \cmidrule(lr){3-6}

   & AL & \textcolor{gray}{92.7 / } 73.8 & \textcolor{gray}{96.9 /} 85.2 & \textcolor{gray}{93.2  /} 73.0 & \textcolor{gray}{97.7 /} 94.5\\

    \noalign{\vskip 1.5pt} 
    \specialrule{1pt}{\aboverulesep}{\belowrulesep}
    \noalign{\vskip 1.5pt} 

    \multirow{2}{*}{\textbf{2}} & AD & \graytext{86.2 /} 70.1 & \graytext{87.6 /} 75.1 & \graytext{84.7 /} 83.4 & \graytext{92.8 /} 81.2 \\

    \cmidrule(lr{0pt}){3-6}

    & AL & \graytext{91.0 /} 68.9 & \graytext{94.8 /} 84.1 & \graytext{91.9 /} 73.9 & \graytext{95.8 /} 94.6 \\

    \noalign{\vskip 1.5pt} 
    \specialrule{1pt}{\aboverulesep}{\belowrulesep}
    \noalign{\vskip 1.5pt} 
    
    \multirow{2}{*}{\textbf{5}} & AD & \graytext{80.1 /} 64.1 & \graytext{78.6 /} 65.4 & \graytext{73.8 /} 69.6 & \graytext{82.5 /} 73.2 \\

    \cmidrule(lr{0pt}){3-6}

    & AL & \graytext{84.7 /} 65.6 & \graytext{81.1 /} 78.9 & \graytext{84.0 /} 65.3 & \graytext{ 86.3 /} 79.9 \\

    \noalign{\vskip 3pt}
    \specialrule{1.5pt}{\aboverulesep}{\belowrulesep}
     \end{tabular}}
    \label{tab:ablation scaling factor}

\end{table}

\subsection{Advanced Attacks} \label{ablation attack}

In our primary experiments in Section \ref{sec:experiments}, we utilized variations of the PGD attack \cite{madry2017towards} for both model training and testing. To further illustrate the adaptability and robustness of our proposed method under diverse adversarial conditions, we also evaluate its performance against a variety of other attack types including Detection attacks and Localization attacks. The Detection attacks consist of black-box attacks \cite{guo2019simple}, FGSM \cite{goodfellow2014explaining}, CAA \cite{Mao2020CompositeAA}, AutoAttack \cite{croce2020reliable}, and $\text{A}^3$ (Adversarial Attack Automation) \cite{liu2022practical}. The localization attacks are adapted from the well-known segmentation attacks in literature, namely SegPGD \cite{gu2022segpgd} and SEA \cite{croce2024robust}. All results are available in Table \ref{tab:advanced attack table}, and the exact details on our adaptation approach for these attacks are available in Appendix \ref{appendix:attack adaptation}. It is important to note that the training process remains consistent, employing only the standard PGD-10 for simplicity and practicality.

\subsection{Pseudo-Anomaly Generation Strategy}

\begin{table}[t]
    \caption{
    An ablation study on our method's performance using various anomaly generation techniques that incorporate masks.}
    
    \resizebox{ \columnwidth}{!}
    {\begin{tabular}{@{}cccccc} 
    
    \specialrule{1.5pt}{\aboverulesep}{\belowrulesep}
    \multirow{2}{*}{\textbf{Pseudo-Anomaly}} & \multirow{2}{*}{\textbf{Task}} & \multicolumn{4}{c}{\textbf{Dataset}} \\
      \cmidrule(lr){3-6}

\textbf{Generator} && \textbf{MVTec AD}& \textbf{VisA}& \textbf{BTAD} & \textbf{BraTS2021}\\ 

    \specialrule{1.5pt}{\aboverulesep}{\belowrulesep} 

    \noalign{\vskip 1.5pt}

   \multirow{2}{*}{CutPaste} & AD & \textcolor{gray}{79.1 / 60.3} & \textcolor{gray}{74.4 / }61.5 & \textcolor{gray}{74.8 / }69.2 & \textcolor{gray}{80.3 / }72.7\\

   \cmidrule(lr){3-6}

   & AL & \textcolor{gray}{83.9 / }60.8 & \textcolor{gray}{85.4 / }73.0 & \textcolor{gray}{82.6 / }61.2 & \textcolor{gray}{91.2 / }82.1\\

    \noalign{\vskip 1.5pt} 
    \specialrule{1pt}{\aboverulesep}{\belowrulesep}
    \noalign{\vskip 1.5pt} 

   \multirow{2}{*}{FPI} & AD & \textcolor{gray}{78.9 / }62.1 & \textcolor{gray}{75.1 / }64.3 & \textcolor{gray}{72.9 / }71.0 & \textcolor{gray}{84.1 / }74.1 \\

   \cmidrule(lr){3-6}

   & AL & \textcolor{gray}{84.3 / }63.8 & \textcolor{gray}{82.9 / }71.8 & \textcolor{gray}{86.3 / }60.5 & \textcolor{gray}{93.3 / }84.8\\

   \noalign{\vskip 1.5pt} 
    \specialrule{1pt}{\aboverulesep}{\belowrulesep}
    \noalign{\vskip 1.5pt} 

   \multirow{2}{*}{NSA} & AD & \textcolor{gray}{81.7 / 66.5} & \textcolor{gray}{83.1 / }65.4 & \textcolor{gray}{77.1 / }73.4 & \textcolor{gray}{86.3 / }73.8\\

   \cmidrule(lr){3-6}

   & AL & \textcolor{gray}{85.1 / }65.0 & \textcolor{gray}{83.3 / }76.7 & \textcolor{gray}{85.4 / }64.9 & \textcolor{gray}{89.4 / }86.8\\

   \noalign{\vskip 1.5pt} 
    \specialrule{1pt}{\aboverulesep}{\belowrulesep}
    \noalign{\vskip 1.5pt} 

   \multirow{2}{*}{GLASS} & AD & \textcolor{gray}{83.2 / }62.0 & \textcolor{gray}{84.0 / }68.7 & \textcolor{gray}{79.3 / }73.9 & \textcolor{gray}{83.1 / }66.1\\

   \cmidrule(lr){3-6}

   & AL & \textcolor{gray}{86.7 / }67.0 & \textcolor{gray}{86.2 / }78.5 & \textcolor{gray}{86.8 / }68.2 & \textcolor{gray}{92.2 / }87.3\\

   \noalign{\vskip 1.5pt} 
    \specialrule{1.5pt}{\aboverulesep}{\belowrulesep}
    \noalign{\vskip 1.5pt} 
    
    \multirow{2}{*}{\textit{Ours}} & AD & \textcolor{gray}{88.1 / }71.1 & \textcolor{gray}{88.5 / }74.3 & \textcolor{gray}{85.3 / }82.1 & \textcolor{gray}{94.3 / }81.0\\

   \cmidrule(lr){3-6}

   & AL & \textcolor{gray}{92.7 / }73.8 & \textcolor{gray}{96.9 /}85.2 & \textcolor{gray}{93.2  /}73.0 & \textcolor{gray}{97.7 /}94.5\\

   \noalign{\vskip 1.5pt} 
    \specialrule{1.5pt}{\aboverulesep}{\belowrulesep}
     \end{tabular}}
    \label{tab:pseudo anomaly table}

\end{table}

Since our method targets both AD and AL using pseudo-anomalies, it is crucial to evaluate our pseudo-anomaly generation against compatible methods that produce localization maps. Thus, methods like MIXUP~\cite{zhang2018mixup}, FakeIt~\cite{mirzaei2022fake}, and Dream-OOD~\cite{du2023dream} are excluded. We compare instead with CutPaste~\cite{li2021cutpaste}, NSA~\cite{schluter2022natural}, GLASS~\cite{chen2024unified}, and FPI~\cite{Tan_2022}, keeping other pipeline components fixed. As shown in Table~\ref{tab:pseudo anomaly table}, our method outperforms these approaches, highlighting its effectiveness for robust anomaly localization.


\section{Conclusion} \label{sec:conclusion}
We introduce PatchGuard, an adversarially robust method for AD and AL. It employs Foreground-Aware Pseudo-Anomaly Generation using segmentation masks. Additionally, we propose a regularization term, motivated by observations and theory, that strengthens robustness against adversarial attacks by increasing the attention degree. PatchGuard outperforms previous methods on AD and AL benchmarks under adversarial settings.


\section*{Acknowledgments}
We are very grateful to Mohammad Javad Maheronnaghsh for his collaboration in the initial stages of this project and to the anonymous reviewers for their helpful discussions and valuable feedback.
{
    \small
    \bibliographystyle{ieeenat_fullname}
    \bibliography{main}

\begin{thebibliography}{95}
\providecommand{\natexlab}[1]{#1}
\providecommand{\url}[1]{\texttt{#1}}
\expandafter\ifx\csname urlstyle\endcsname\relax
  \providecommand{\doi}[1]{doi: #1}\else
  \providecommand{\doi}{doi: \begingroup \urlstyle{rm}\Url}\fi

\bibitem[Akbiyik(2019)]{Akbiyik2019DataAI}
M.~Eren Akbiyik.
\newblock Data augmentation in training cnns: Injecting noise to images.
\newblock \emph{ArXiv}, abs/2307.06855, 2019.

\bibitem[Akhtar and Mian(2018)]{akhtar2018threat}
Naveed Akhtar and Ajmal Mian.
\newblock Threat of adversarial attacks on deep learning in computer vision: A survey.
\newblock \emph{Ieee Access}, 6:\penalty0 14410--14430, 2018.

\bibitem[Aota et~al.(2023)Aota, Tong, and Okatani]{Aota_2023_WACV}
Toshimichi Aota, Lloyd Teh~Tzer Tong, and Takayuki Okatani.
\newblock Zero-shot versus many-shot: Unsupervised texture anomaly detection.
\newblock In \emph{Proceedings of the IEEE/CVF Winter Conference on Applications of Computer Vision (WACV)}, pages 5564--5572, 2023.

\bibitem[Azizmalayeri et~al.(2022)Azizmalayeri, Soltani~Moakhar, Zarei, Zohrabi, Manzuri, and Rohban]{azizmalayeri2022your}
Mohammad Azizmalayeri, Arshia Soltani~Moakhar, Arman Zarei, Reihaneh Zohrabi, Mohammad Manzuri, and Mohammad~Hossein Rohban.
\newblock Your out-of-distribution detection method is not robust!
\newblock \emph{Advances in Neural Information Processing Systems}, 35:\penalty0 4887--4901, 2022.

\bibitem[Baid et~al.(2021)Baid, Ghodasara, Mohan, Bilello, Calabrese, Colak, Farahani, Kalpathy-Cramer, Kitamura, Pati, et~al.]{baid2021rsna}
Ujjwal Baid, Satyam Ghodasara, Suyash Mohan, Michel Bilello, Evan Calabrese, Errol Colak, Keyvan Farahani, Jayashree Kalpathy-Cramer, Felipe~C Kitamura, Sarthak Pati, et~al.
\newblock The rsna-asnr-miccai brats 2021 benchmark on brain tumor segmentation and radiogenomic classification.
\newblock \emph{arXiv preprint arXiv:2107.02314}, 2021.

\bibitem[Bendale and Boult(2015)]{bendale2015towards}
Abhijit Bendale and Terrance Boult.
\newblock Towards open world recognition.
\newblock In \emph{Proceedings of the IEEE Conference on Computer Vision and Pattern Recognition (CVPR)}, pages 1893--1902, 2015.

\bibitem[Bergmann et~al.(2019)Bergmann, Fauser, Sattlegger, and Steger]{bergmann2019mvtec}
Paul Bergmann, Michael Fauser, David Sattlegger, and Carsten Steger.
\newblock Mvtec ad--a comprehensive real-world dataset for unsupervised anomaly detection.
\newblock In \emph{Proceedings of the IEEE/CVF conference on computer vision and pattern recognition}, pages 9592--9600, 2019.

\bibitem[B{\'e}thune et~al.(2023)B{\'e}thune, Novello, Boissin, Coiffier, Serrurier, Vincenot, and Troya-Galvis]{bethune2023robust}
Louis B{\'e}thune, Paul Novello, Thibaut Boissin, Guillaume Coiffier, Mathieu Serrurier, Quentin Vincenot, and Andres Troya-Galvis.
\newblock Robust one-class classification with signed distance function using 1-lipschitz neural networks.
\newblock \emph{arXiv preprint arXiv:2303.01978}, 2023.

\bibitem[Caron et~al.(2020)Caron, Misra, Mairal, Goyal, Bojanowski, and Joulin]{caron2020unsupervised}
Mathilde Caron, Ishan Misra, Julien Mairal, Priya Goyal, Piotr Bojanowski, and Armand Joulin.
\newblock Unsupervised learning of visual features by contrasting cluster assignments.
\newblock \emph{Advances in neural information processing systems}, 33:\penalty0 9912--9924, 2020.

\bibitem[Chen et~al.(2021{\natexlab{a}})Chen, Xie, Lin, Qiao, Zhou, Tan, Zhang, and Ma]{DBLP:conf/ijcai/ChenXLQZTZM21}
Chengwei Chen, Yuan Xie, Shaohui Lin, Ruizhi Qiao, Jian Zhou, Xin Tan, Yi Zhang, and Lizhuang Ma.
\newblock Novelty detection via contrastive learning with negative data augmentation.
\newblock In \emph{Proceedings of the Thirtieth International Joint Conference on Artificial Intelligence, {IJCAI} 2021, Virtual Event / Montreal, Canada, 19-27 August 2021}, pages 606--614. ijcai.org, 2021{\natexlab{a}}.

\bibitem[Chen et~al.(2020{\natexlab{a}})Chen, Li, Wu, Liang, and Jha]{chen2020robust}
Jiefeng Chen, Yixuan Li, Xi Wu, Yingyu Liang, and Somesh Jha.
\newblock Robust out-of-distribution detection for neural networks.
\newblock \emph{arXiv preprint arXiv:2003.09711}, 2020{\natexlab{a}}.

\bibitem[Chen et~al.(2021{\natexlab{b}})Chen, Li, Wu, Liang, and Jha]{chen2021atom}
Jiefeng Chen, Yixuan Li, Xi Wu, Yingyu Liang, and Somesh Jha.
\newblock Atom: Robustifying out-of-distribution detection using outlier mining.
\newblock In \emph{Machine Learning and Knowledge Discovery in Databases. Research Track: European Conference, ECML PKDD 2021, Bilbao, Spain, September 13--17, 2021, Proceedings, Part III 21}, pages 430--445. Springer, 2021{\natexlab{b}}.

\bibitem[Chen et~al.(2024)Chen, Luo, Lv, and Zhang]{chen2024unified}
Qiyu Chen, Huiyuan Luo, Chengkan Lv, and Zhengtao Zhang.
\newblock A unified anomaly synthesis strategy with gradient ascent for industrial anomaly detection and localization.
\newblock \emph{arXiv preprint arXiv:2407.09359}, 2024.

\bibitem[Chen et~al.(2020{\natexlab{b}})Chen, Kornblith, Norouzi, and Hinton]{chen2020simple}
Ting Chen, Simon Kornblith, Mohammad Norouzi, and Geoffrey Hinton.
\newblock A simple framework for contrastive learning of visual representations.
\newblock In \emph{International conference on machine learning}, pages 1597--1607. PMLR, 2020{\natexlab{b}}.

\bibitem[Chen and He(2021)]{chen2021exploring}
Xinlei Chen and Kaiming He.
\newblock Exploring simple siamese representation learning.
\newblock In \emph{Proceedings of the IEEE/CVF conference on computer vision and pattern recognition}, pages 15750--15758, 2021.

\bibitem[Cohen and Avidan(2021)]{cohen2021transformaly}
Matan~Jacob Cohen and Shai Avidan.
\newblock Transformaly--two (feature spaces) are better than one.
\newblock \emph{arXiv preprint arXiv:2112.04185}, 2021.

\bibitem[Croce and Hein(2020)]{croce2020reliable}
Francesco Croce and Matthias Hein.
\newblock Reliable evaluation of adversarial robustness with an ensemble of diverse parameter-free attacks.
\newblock In \emph{International conference on machine learning}, pages 2206--2216. PMLR, 2020.

\bibitem[Croce et~al.(2024)Croce, Singh, and Hein]{croce2024robust}
Francesco Croce, Naman~D Singh, and Matthias Hein.
\newblock Towards reliable evaluation and fast training of robust semantic segmentation models.
\newblock In \emph{ECCV}, 2024.

\bibitem[Deng et~al.(2009)Deng, Dong, Socher, Li, Li, and Fei-Fei]{deng2009imagenet}
Jia Deng, Wei Dong, Richard Socher, Li-Jia Li, Kai Li, and Li Fei-Fei.
\newblock Imagenet: A large-scale hierarchical image database.
\newblock In \emph{2009 IEEE Conference on Computer Vision and Pattern Recognition}, pages 248--255. IEEE, 2009.

\bibitem[DeVries and Taylor(2017)]{devries2017cutout}
Terrance DeVries and Graham~W Taylor.
\newblock Improved regularization of convolutional neural networks with cutout.
\newblock \emph{arXiv preprint arXiv:1708.04552}, 2017.

\bibitem[Dosovitskiy et~al.(2020)Dosovitskiy, Beyer, Kolesnikov, Weissenborn, Zhai, Unterthiner, Dehghani, Minderer, Heigold, Gelly, et~al.]{dosovitskiy2020image}
Alexey Dosovitskiy, Lucas Beyer, Alexander Kolesnikov, Dirk Weissenborn, Xiaohua Zhai, Thomas Unterthiner, Mostafa Dehghani, Matthias Minderer, Georg Heigold, Sylvain Gelly, et~al.
\newblock An image is worth 16x16 words: Transformers for image recognition at scale.
\newblock \emph{arXiv preprint arXiv:2010.11929}, 2020.

\bibitem[Du et~al.(2023)Du, Sun, Zhu, and Li]{du2023dream}
Xuefeng Du, Yiyou Sun, Xiaojin Zhu, and Yixuan Li.
\newblock Dream the impossible: Outlier imagination with diffusion models.
\newblock \emph{arXiv preprint arXiv:2309.13415}, 2023.

\bibitem[Feinman et~al.(2017)Feinman, Curtin, Shintre, and Gardner]{feinman2017detecting}
Reuben Feinman, Ryan~R Curtin, Saurabh Shintre, and Andrew~B Gardner.
\newblock Detecting adversarial samples from artifacts.
\newblock \emph{arXiv preprint arXiv:1703.00410}, 2017.

\bibitem[Fernando et~al.(2021)Fernando, Gammulle, Denman, Sridharan, and Fookes]{fernando2021deep}
Tharindu Fernando, Harshala Gammulle, Simon Denman, Sridha Sridharan, and Clinton Fookes.
\newblock Deep learning for medical anomaly detection--a survey.
\newblock \emph{ACM Computing Surveys (CSUR)}, 54\penalty0 (7):\penalty0 1--37, 2021.

\bibitem[Ghiasi et~al.(2020)Ghiasi, Cui, Srinivas, Qian, Lin, Cubuk, Le, and Zoph]{ghiasi2020simple}
Golnaz Ghiasi, Yin Cui, Aravind Srinivas, Rui Qian, Tsung-Yi Lin, Ekin~D Cubuk, Quoc~V Le, and Barret Zoph.
\newblock Simple copy-paste is a strong data augmentation method for instance segmentation.
\newblock \emph{arXiv preprint arXiv:2012.07177}, 2020.

\bibitem[Goodfellow et~al.(2014)Goodfellow, Shlens, and Szegedy]{goodfellow2014explaining}
Ian~J Goodfellow, Jonathon Shlens, and Christian Szegedy.
\newblock Explaining and harnessing adversarial examples.
\newblock \emph{arXiv preprint arXiv:1412.6572}, 2014.

\bibitem[Goodge et~al.(2021)Goodge, Hooi, Ng, and Ng]{goodge2021robustness}
Adam Goodge, Bryan Hooi, See~Kiong Ng, and Wee~Siong Ng.
\newblock Robustness of autoencoders for anomaly detection under adversarial impact.
\newblock In \emph{Proceedings of the Twenty-Ninth International Conference on International Joint Conferences on Artificial Intelligence}, pages 1244--1250, 2021.

\bibitem[Grill et~al.(2020)Grill, Strub, Altch{\'e}, Tallec, Richemond, Buchatskaya, Doersch, Avila~Pires, Guo, Gheshlaghi~Azar, et~al.]{grill2020bootstrap}
Jean-Bastien Grill, Florian Strub, Florent Altch{\'e}, Corentin Tallec, Pierre Richemond, Elena Buchatskaya, Carl Doersch, Bernardo Avila~Pires, Zhaohan Guo, Mohammad Gheshlaghi~Azar, et~al.
\newblock Bootstrap your own latent-a new approach to self-supervised learning.
\newblock \emph{Advances in neural information processing systems}, 33:\penalty0 21271--21284, 2020.

\bibitem[Gu et~al.(2022)Gu, Zhao, Tresp, and Torr]{gu2022segpgd}
Jindong Gu, Hengshuang Zhao, Volker Tresp, and Philip~HS Torr.
\newblock Segpgd: An effective and efficient adversarial attack for evaluating and boosting segmentation robustness.
\newblock In \emph{European Conference on Computer Vision}, pages 308--325. Springer, 2022.

\bibitem[Guo et~al.(2019)Guo, Gardner, You, Wilson, and Weinberger]{guo2019simple}
Chuan Guo, Jacob Gardner, Yurong You, Andrew~Gordon Wilson, and Kilian Weinberger.
\newblock Simple black-box adversarial attacks.
\newblock In \emph{International Conference on Machine Learning}, pages 2484--2493. PMLR, 2019.

\bibitem[Guo et~al.(2024)Guo, Jia, Zhang, Li, et~al.]{guo2024recontrast}
Jia Guo, Lize Jia, Weihang Zhang, Huiqi Li, et~al.
\newblock Recontrast: Domain-specific anomaly detection via contrastive reconstruction.
\newblock \emph{Advances in Neural Information Processing Systems}, 36, 2024.

\bibitem[He et~al.(2016)He, Zhang, Ren, and Sun]{resnet}
Kaiming He, Xiangyu Zhang, Shaoqing Ren, and Jian Sun.
\newblock Deep residual learning for image recognition.
\newblock In \emph{2016 IEEE Conference on Computer Vision and Pattern Recognition (CVPR)}, pages 770--778, 2016.

\bibitem[He et~al.(2020)He, Fan, Wu, Xie, and Girshick]{he2020momentum}
Kaiming He, Haoqi Fan, Yuxin Wu, Saining Xie, and Ross Girshick.
\newblock Momentum contrast for unsupervised visual representation learning.
\newblock In \emph{Proceedings of the IEEE/CVF conference on computer vision and pattern recognition}, pages 9729--9738, 2020.

\bibitem[Hendrycks et~al.(2018)Hendrycks, Mazeika, and Dietterich]{hendrycks2018deep}
Dan Hendrycks, Mantas Mazeika, and Thomas Dietterich.
\newblock Deep anomaly detection with outlier exposure.
\newblock \emph{arXiv preprint arXiv:1812.04606}, 2018.

\bibitem[Hongyi~Zhang(2018)]{zhang2018mixup}
Yann N.~Dauphin Hongyi~Zhang, Moustapha~Cisse.
\newblock mixup: Beyond empirical risk minimization.
\newblock \emph{International Conference on Learning Representations}, 2018.

\bibitem[Hossein~Mirzaei(2024)]{mirzaei2022scan}
Mohammad Hossein~Rohban Hossein~Mirzaei, ...
\newblock Scanning trojaned models using out-of-distribution samples.
\newblock In \emph{Advances in Neural Information Processing Systems}, 2024.

\bibitem[Huang et~al.(2025)Huang, Li, Li, Wu, Zhang, Yuan, and Zhang]{huang2025real}
Jiande Huang, Xingxing Li, Xin Li, Jiaqi Wu, Keke Zhang, Yongqiang Yuan, and Wei Zhang.
\newblock Real-time outlier detection of satellite orbit and clock products using reverse error estimation.
\newblock \emph{GPS Solutions}, 29\penalty0 (1):\penalty0 2, 2025.

\bibitem[Jeffrey et~al.(2023)Jeffrey, Tan, and Villar]{jeffrey2023review}
Nicholas Jeffrey, Qing Tan, and Jos{\'e}~R Villar.
\newblock A review of anomaly detection strategies to detect threats to cyber-physical systems.
\newblock \emph{Electronics}, 12\penalty0 (15):\penalty0 3283, 2023.

\bibitem[Jeong et~al.(2023)Jeong, Zou, Kim, Zhang, Ravichandran, and Dabeer]{jeong2023winclip}
Jongheon Jeong, Yang Zou, Taewan Kim, Dongqing Zhang, Avinash Ravichandran, and Onkar Dabeer.
\newblock Winclip: Zero-/few-shot anomaly classification and segmentation.
\newblock In \emph{Proceedings of the IEEE/CVF Conference on Computer Vision and Pattern Recognition}, pages 19606--19616, 2023.

\bibitem[Jezek et~al.(2021)Jezek, Jonak, Burget, Dvorak, and Skotak]{9631567}
Stepan Jezek, Martin Jonak, Radim Burget, Pavel Dvorak, and Milos Skotak.
\newblock Deep learning-based defect detection of metal parts: evaluating current methods in complex conditions.
\newblock In \emph{2021 13th International Congress on Ultra Modern Telecommunications and Control Systems and Workshops (ICUMT)}, pages 66--71, 2021.

\bibitem[Kalantidis et~al.(2020)Kalantidis, Sariyildiz, Pion, Weinzaepfel, and Larlus]{kalantidis2020hard}
Yannis Kalantidis, Mert~Bulent Sariyildiz, Noe Pion, Philippe Weinzaepfel, and Diane Larlus.
\newblock Hard negative mixing for contrastive learning.
\newblock \emph{Advances in Neural Information Processing Systems}, 33:\penalty0 21798--21809, 2020.

\bibitem[Kitamura(2018)]{kitamura2018headct}
F.C. Kitamura.
\newblock Head ct - hemorrhage, 2018.

\bibitem[Lee et~al.(2022)Lee, Kim, Mok, and Yoon]{lee2022anti}
Jungbeom Lee, Eunji Kim, Jisoo Mok, and Sungroh Yoon.
\newblock Anti-adversarially manipulated attributions for weakly supervised semantic segmentation and object localization.
\newblock \emph{IEEE transactions on pattern analysis and machine intelligence}, 46\penalty0 (3):\penalty0 1618--1634, 2022.

\bibitem[Li et~al.(2021)Li, Sohn, Yoon, and Pfister]{li2021cutpaste}
Chun-Liang Li, Kihyuk Sohn, Jinsung Yoon, and Tomas Pfister.
\newblock Cutpaste: Self-supervised learning for anomaly detection and localization.
\newblock In \emph{Proceedings of the IEEE/CVF Conference on Computer Vision and Pattern Recognition}, pages 9664--9674, 2021.

\bibitem[Li et~al.(2024)Li, Huang, Xue, and Zhou]{li2024musc}
Xurui Li, Ziming Huang, Feng Xue, and Yu Zhou.
\newblock Musc: Zero-shot industrial anomaly classification and segmentation with mutual scoring of the unlabeled images.
\newblock \emph{arXiv preprint arXiv:2401.16753}, 2024.

\bibitem[Lin(2017)]{lin2017focal}
T Lin.
\newblock Focal loss for dense object detection.
\newblock \emph{arXiv preprint arXiv:1708.02002}, 2017.

\bibitem[Liu et~al.(2024)Liu, Xie, Wang, Li, Wang, Zheng, and Jin]{liu2024deep}
Jiaqi Liu, Guoyang Xie, Jinbao Wang, Shangnian Li, Chengjie Wang, Feng Zheng, and Yaochu Jin.
\newblock Deep industrial image anomaly detection: A survey.
\newblock \emph{Machine Intelligence Research}, 21\penalty0 (1):\penalty0 104--135, 2024.

\bibitem[Liu et~al.(2022)Liu, Cheng, Gao, Liu, Zhang, and Song]{liu2022practical}
Ye Liu, Yaya Cheng, Lianli Gao, Xianglong Liu, Qilong Zhang, and Jingkuan Song.
\newblock Practical evaluation of adversarial robustness via adaptive auto attack, 2022.

\bibitem[Liu et~al.(2023)Liu, Zhou, Xu, and Wang]{liu2023simplenet}
Zhikang Liu, Yiming Zhou, Yuansheng Xu, and Zilei Wang.
\newblock Simplenet: A simple network for image anomaly detection and localization.
\newblock In \emph{Proceedings of the IEEE/CVF Conference on Computer Vision and Pattern Recognition}, pages 20402--20411, 2023.

\bibitem[Lo et~al.(2022)Lo, Oza, and Patel]{lo2022adversarially}
Shao-Yuan Lo, Poojan Oza, and Vishal~M Patel.
\newblock Adversarially robust one-class novelty detection.
\newblock \emph{IEEE Transactions on Pattern Analysis and Machine Intelligence}, 2022.

\bibitem[Loshchilov and Hutter(2019)]{loshchilov2018decoupled}
Ilya Loshchilov and Frank Hutter.
\newblock Decoupled weight decay regularization.
\newblock In \emph{International Conference on Learning Representations}, 2019.

\bibitem[Luc et~al.(2016)Luc, Couprie, Chintala, and Verbeek]{luc2016semantic}
Pauline Luc, Camille Couprie, Soumith Chintala, and Jakob Verbeek.
\newblock Semantic segmentation using adversarial networks.
\newblock \emph{arXiv preprint arXiv:1611.08408}, 2016.

\bibitem[Luo et~al.(2021)Luo, Xiao, Cheng, Peng, and Yao]{luo2021deep}
Yuan Luo, Ya Xiao, Long Cheng, Guojun Peng, and Danfeng Yao.
\newblock Deep learning-based anomaly detection in cyber-physical systems: Progress and opportunities.
\newblock \emph{ACM Computing Surveys (CSUR)}, 54\penalty0 (5):\penalty0 1--36, 2021.

\bibitem[Madry et~al.(2017)Madry, Makelov, Schmidt, Tsipras, and Vladu]{madry2017towards}
Aleksander Madry, Aleksandar Makelov, Ludwig Schmidt, Dimitris Tsipras, and Adrian Vladu.
\newblock Towards deep learning models resistant to adversarial attacks.
\newblock \emph{arXiv preprint arXiv:1706.06083}, 2017.

\bibitem[Mao et~al.(2020)Mao, Chen, Wang, Su, He, and Xue]{Mao2020CompositeAA}
Xiaofeng Mao, Yuefeng Chen, Shuhui Wang, Hang Su, Yuan He, and Hui Xue.
\newblock Composite adversarial attacks.
\newblock \emph{ArXiv}, abs/2012.05434, 2020.

\bibitem[Mirzaei et~al.(2022)Mirzaei, Salehi, Shahabi, Gavves, Snoek, Sabokrou, and Rohban]{mirzaei2022fake}
Hossein Mirzaei, Mohammadreza Salehi, Sajjad Shahabi, Efstratios Gavves, Cees~GM Snoek, Mohammad Sabokrou, and Mohammad~Hossein Rohban.
\newblock Fake it till you make it: Near-distribution novelty detection by score-based generative models.
\newblock \emph{arXiv preprint arXiv:2205.14297}, 2022.

\bibitem[Mirzaei et~al.(2024{\natexlab{a}})Mirzaei, Jafari, Dehbashi, Ansari, Ghobadi, Hadi, Moakhar, Azizmalayeri, Baghshah, and Rohban]{mirzaeirodeo}
Hossein Mirzaei, Mohammad Jafari, Hamid~Reza Dehbashi, Ali Ansari, Sepehr Ghobadi, Masoud Hadi, Arshia~Soltani Moakhar, Mohammad Azizmalayeri, Mahdieh~Soleymani Baghshah, and Mohammad~Hossein Rohban.
\newblock Rodeo: Robust outlier detection via exposing adaptive out-of-distribution samples.
\newblock In \emph{Forty-first International Conference on Machine Learning}, 2024{\natexlab{a}}.

\bibitem[Mirzaei et~al.(2024{\natexlab{b}})Mirzaei, Jafari, Dehbashi, Taghavi, Sabokrou, and Rohban]{mirzaei2024killing}
Hossein Mirzaei, Mohammad Jafari, Hamid~Reza Dehbashi, Zeinab~Sadat Taghavi, Mohammad Sabokrou, and Mohammad~Hossein Rohban.
\newblock Killing it with zero-shot: Adversarially robust novelty detection.
\newblock In \emph{ICASSP 2024-2024 IEEE International Conference on Acoustics, Speech and Signal Processing (ICASSP)}, pages 7415--7419. IEEE, 2024{\natexlab{b}}.

\bibitem[Mirzaei et~al.(2024{\natexlab{c}})Mirzaei, Nafez, Jafari, Soltani, Azizmalayeri, Habibi, Sabokrou, and Rohban]{mirzaei2024universal}
Hossein Mirzaei, Mojtaba Nafez, Mohammad Jafari, Mohammad~Bagher Soltani, Mohammad Azizmalayeri, Jafar Habibi, Mohammad Sabokrou, and Mohammad~Hossein Rohban.
\newblock Universal novelty detection through adaptive contrastive learning.
\newblock In \emph{Proceedings of the IEEE/CVF Conference on Computer Vision and Pattern Recognition}, pages 22914--22923, 2024{\natexlab{c}}.

\bibitem[Mirzaei et~al.(2025{\natexlab{a}})Mirzaei, Nafez, Habibi, Sabokrou, and Rohban]{mirzaei2025mitigatingspuriousnegativepairs}
Hossein Mirzaei, Mojtaba Nafez, Jafar Habibi, Mohammad Sabokrou, and Mohammad~Hossein Rohban.
\newblock Mitigating spurious negative pairs for robust industrial anomaly detection, 2025{\natexlab{a}}.

\bibitem[Mirzaei et~al.(2025{\natexlab{b}})Mirzaei, Nafez, Madadi, Maleki, Hajialilue, Taghavi, Rezaee, Ansari, Nia, Shamsaie, Salehi, Mathis, Baghshah, Sabokrou, and Rohban]{mirzaei2025contrastiveteacherstudentframeworknovelty}
Hossein Mirzaei, Mojtaba Nafez, Moein Madadi, Arad Maleki, Mahdi Hajialilue, Zeinab~Sadat Taghavi, Sepehr Rezaee, Ali Ansari, Bahar~Dibaei Nia, Kian Shamsaie, Mohammadreza Salehi, Mackenzie~W. Mathis, Mahdieh~Soleymani Baghshah, Mohammad Sabokrou, and Mohammad~Hossein Rohban.
\newblock A contrastive teacher-student framework for novelty detection under style shifts, 2025{\natexlab{b}}.

\bibitem[Mishra et~al.(2021)Mishra, Verk, Fornasier, Piciarelli, and Foresti]{mishra21-vt-adl}
Pankaj Mishra, Riccardo Verk, Daniele Fornasier, Claudio Piciarelli, and Gian~Luca Foresti.
\newblock {VT-ADL}: A vision transformer network for image anomaly detection and localization.
\newblock In \emph{30th IEEE/IES International Symposium on Industrial Electronics (ISIE)}, 2021.

\bibitem[Mousakhan et~al.(2023)Mousakhan, Brox, and Tayyub]{mousakhan2023anomaly}
Arian Mousakhan, Thomas Brox, and Jawad Tayyub.
\newblock Anomaly detection with conditioned denoising diffusion models.
\newblock \emph{arXiv preprint arXiv:2305.15956}, 2023.

\bibitem[Noohdani et~al.(2024)Noohdani, Hosseini, Parast, Araghi, and Baghshah]{noohdani2024decompose}
Fahimeh~Hosseini Noohdani, Parsa Hosseini, Aryan~Yazdan Parast, Hamidreza~Yaghoubi Araghi, and Mahdieh~Soleymani Baghshah.
\newblock Decompose-and-compose: A compositional approach to mitigating spurious correlation.
\newblock In \emph{Proceedings of the IEEE/CVF Conference on Computer Vision and Pattern Recognition}, pages 27662--27671, 2024.

\bibitem[Perera et~al.(2021)Perera, Oza, and Patel]{perera2021one}
Pramuditha Perera, Poojan Oza, and Vishal~M Patel.
\newblock One-class classification: A survey.
\newblock \emph{arXiv preprint arXiv:2101.03064}, 2021.

\bibitem[Ronneberger et~al.(2015)Ronneberger, Fischer, and Brox]{ronneberger2015u}
Olaf Ronneberger, Philipp Fischer, and Thomas Brox.
\newblock U-net: Convolutional networks for biomedical image segmentation.
\newblock In \emph{Medical image computing and computer-assisted intervention--MICCAI 2015: 18th international conference, Munich, Germany, October 5-9, 2015, proceedings, part III 18}, pages 234--241. Springer, 2015.

\bibitem[Roth et~al.(2021)Roth, Pemula, Zepeda, Schölkopf, Brox, and Gehler]{roth2021total}
Karsten Roth, Latha Pemula, Joaquin Zepeda, Bernhard Schölkopf, Thomas Brox, and Peter Gehler.
\newblock Towards total recall in industrial anomaly detection, 2021.

\bibitem[Salehi et~al.(2021)Salehi, Mirzaei, Hendrycks, Li, Rohban, and Sabokrou]{salehi2021unified}
Mohammadreza Salehi, Hossein Mirzaei, Dan Hendrycks, Yixuan Li, Mohammad~Hossein Rohban, and Mohammad Sabokrou.
\newblock A unified survey on anomaly, novelty, open-set, and out-of-distribution detection: Solutions and future challenges.
\newblock \emph{arXiv preprint arXiv:2110.14051}, 2021.

\bibitem[Schl{\"u}ter et~al.(2022)Schl{\"u}ter, Tan, Hou, and Kainz]{schluter2022natural}
Hannah~M Schl{\"u}ter, Jeremy Tan, Benjamin Hou, and Bernhard Kainz.
\newblock Natural synthetic anomalies for self-supervised anomaly detection and localization.
\newblock In \emph{European Conference on Computer Vision}, pages 474--489. Springer, 2022.

\bibitem[Selvaraju et~al.(2017)Selvaraju, Cogswell, Das, Vedantam, Parikh, and Batra]{gradcam}
Ramprasaath~R Selvaraju, Michael Cogswell, Abhishek Das, Ramakrishna Vedantam, Devi Parikh, and Dhruv Batra.
\newblock Grad-cam: Visual explanations from deep networks via gradient-based localization.
\newblock In \emph{Proceedings of the IEEE international conference on computer vision}, pages 618--626, 2017.

\bibitem[Shao et~al.(2020)Shao, Perera, Yuen, and Patel]{shao2020open}
Rui Shao, Pramuditha Perera, Pong~C Yuen, and Vishal~M Patel.
\newblock Open-set adversarial defense.
\newblock In \emph{Computer Vision--ECCV 2020: 16th European Conference, Glasgow, UK, August 23--28, 2020, Proceedings, Part XVII 16}, pages 682--698. Springer, 2020.

\bibitem[Sinha et~al.(2021)Sinha, Ayush, Song, Uzkent, Jin, and Ermon]{sinha2021negative}
Abhishek Sinha, Kumar Ayush, Jiaming Song, Burak Uzkent, Hongxia Jin, and Stefano Ermon.
\newblock Negative data augmentation.
\newblock In \emph{International Conference on Learning Representations}, 2021.

\bibitem[Sohn et~al.(2020)Sohn, Li, Yoon, Jin, and Pfister]{sohn2020learning}
Kihyuk Sohn, Chun-Liang Li, Jinsung Yoon, Minho Jin, and Tomas Pfister.
\newblock Learning and evaluating representations for deep one-class classification.
\newblock \emph{arXiv preprint arXiv:2011.02578}, 2020.

\bibitem[Str{\"a}ter et~al.(2024)Str{\"a}ter, Salehi, Gavves, Snoek, and Asano]{strater2024generalad}
Luc~PJ Str{\"a}ter, Mohammadreza Salehi, Efstratios Gavves, Cees~GM Snoek, and Yuki~M Asano.
\newblock Generalad: Anomaly detection across domains by attending to distorted features.
\newblock \emph{arXiv preprint arXiv:2407.12427}, 2024.

\bibitem[Szegedy et~al.(2013)Szegedy, Zaremba, Sutskever, Bruna, Erhan, Goodfellow, and Fergus]{szegedy2013intriguing}
Christian Szegedy, Wojciech Zaremba, Ilya Sutskever, Joan Bruna, Dumitru Erhan, Ian Goodfellow, and Rob Fergus.
\newblock Intriguing properties of neural networks.
\newblock \emph{arXiv preprint arXiv:1312.6199}, 2013.

\bibitem[Tack et~al.(2020)Tack, Mo, Jeong, and Shin]{tack2020csi}
Jihoon Tack, Sangwoo Mo, Jongheon Jeong, and Jinwoo Shin.
\newblock Csi: Novelty detection via contrastive learning on distributionally shifted instances.
\newblock \emph{Advances in neural information processing systems}, 33:\penalty0 11839--11852, 2020.

\bibitem[Tan et~al.(2022)Tan, Hou, Batten, Qiu, and Kainz]{Tan_2022}
Jeremy Tan, Benjamin Hou, James Batten, Huaqi Qiu, and Bernhard Kainz.
\newblock Detecting outliers with foreign patch interpolation.
\newblock \emph{Machine Learning for Biomedical Imaging}, 1\penalty0 (April 2022):\penalty0 1–27, 2022.

\bibitem[Tschuchnig and Gadermayr(2022)]{tschuchnig2022anomaly}
Maximilian~E Tschuchnig and Michael Gadermayr.
\newblock Anomaly detection in medical imaging-a mini review.
\newblock In \emph{Data Science--Analytics and Applications: Proceedings of the 4th International Data Science Conference--iDSC2021}, pages 33--38. Springer, 2022.

\bibitem[Wang et~al.(2019)Wang, Gao, and Li]{wang2019weakly}
Qi Wang, Junyu Gao, and Xuelong Li.
\newblock Weakly supervised adversarial domain adaptation for semantic segmentation in urban scenes.
\newblock \emph{IEEE Transactions on Image Processing}, 28\penalty0 (9):\penalty0 4376--4386, 2019.

\bibitem[Wang et~al.(2021)Wang, Wang, Zhou, Deng, Zhao, Wang, and Guo]{wang2021anomaly}
Weiping Wang, Zhaorong Wang, Zhanfan Zhou, Haixia Deng, Weiliang Zhao, Chunyang Wang, and Yongzhen Guo.
\newblock Anomaly detection of industrial control systems based on transfer learning.
\newblock \emph{Tsinghua Science and Technology}, 26\penalty0 (6):\penalty0 821--832, 2021.

\bibitem[Xing et~al.(2022)Xing, Song, and Cheng]{xing2022artificially}
Yue Xing, Qifan Song, and Guang Cheng.
\newblock Why do artificially generated data help adversarial robustness.
\newblock \emph{Advances in Neural Information Processing Systems}, 35:\penalty0 954--966, 2022.

\bibitem[Xu et~al.(2019)Xu, Ma, Liu, Deb, Liu, Tang, and Jain]{xu2019adversarialattacksdefensesimages}
Han Xu, Yao Ma, Haochen Liu, Debayan Deb, Hui Liu, Jiliang Tang, and Anil~K. Jain.
\newblock Adversarial attacks and defenses in images, graphs and text: A review, 2019.

\bibitem[Yang et~al.(2024)Yang, Zhou, Li, and Liu]{yang2024generalized}
Jingkang Yang, Kaiyang Zhou, Yixuan Li, and Ziwei Liu.
\newblock Generalized out-of-distribution detection: A survey.
\newblock \emph{International Journal of Computer Vision}, pages 1--28, 2024.

\bibitem[Yao et~al.(2024)Yao, Liu, Wang, Yin, Yan, Hong, and Zuo]{yao2024glad}
Hang Yao, Ming Liu, Haolin Wang, Zhicun Yin, Zifei Yan, Xiaopeng Hong, and Wangmeng Zuo.
\newblock Glad: Towards better reconstruction with global and local adaptive diffusion models for unsupervised anomaly detection.
\newblock \emph{arXiv preprint arXiv:2406.07487}, 2024.

\bibitem[You et~al.(2022)You, Cui, Shen, Yang, Lu, Zheng, and Le]{you2022unified}
Zhiyuan You, Lei Cui, Yujun Shen, Kai Yang, Xin Lu, Yu Zheng, and Xinyi Le.
\newblock A unified model for multi-class anomaly detection.
\newblock \emph{Advances in Neural Information Processing Systems}, 35:\penalty0 4571--4584, 2022.

\bibitem[Yuan et~al.(2019)Yuan, He, Zhu, and Li]{yuan2019adversarial}
Xiaoyong Yuan, Pan He, Qile Zhu, and Xiaolin Li.
\newblock Adversarial examples: Attacks and defenses for deep learning.
\newblock \emph{IEEE transactions on neural networks and learning systems}, 30\penalty0 (9):\penalty0 2805--2824, 2019.

\bibitem[Yun et~al.(2019)Yun, Han, Oh, Chun, Choe, and Yoo]{yun2019cutmix}
Sangdoo Yun, Dongyoon Han, Seong~Joon Oh, Sanghyuk Chun, Junsuk Choe, and Youngjoon Yoo.
\newblock Cutmix: Regularization strategy to train strong classifiers with localizable features, 2019.

\bibitem[Zagoruyko and Komodakis(2017)]{zagoruyko2017wide}
Sergey Zagoruyko and Nikos Komodakis.
\newblock Wide residual networks, 2017.

\bibitem[Zaheer et~al.(2021)Zaheer, Guruganesh, Dubey, Ainslie, Alberti, Ontanon, Pham, Ravula, Wang, Yang, and Ahmed]{zaheer2021bigbirdtransformerslonger}
Manzil Zaheer, Guru Guruganesh, Avinava Dubey, Joshua Ainslie, Chris Alberti, Santiago Ontanon, Philip Pham, Anirudh Ravula, Qifan Wang, Li Yang, and Amr Ahmed.
\newblock Big bird: Transformers for longer sequences, 2021.

\bibitem[Zavrtanik et~al.(2021)Zavrtanik, Kristan, and Sko{\v{c}}aj]{zavrtanik2021draem}
Vitjan Zavrtanik, Matej Kristan, and Danijel Sko{\v{c}}aj.
\newblock Draem-a discriminatively trained reconstruction embedding for surface anomaly detection.
\newblock In \emph{Proceedings of the IEEE/CVF International Conference on Computer Vision}, pages 8330--8339, 2021.

\bibitem[Zavrtanik et~al.(2022)Zavrtanik, Kristan, and Skočaj]{zavrtanik2022dsrdualsubspace}
Vitjan Zavrtanik, Matej Kristan, and Danijel Skočaj.
\newblock Dsr -- a dual subspace re-projection network for surface anomaly detection, 2022.

\bibitem[Zhang et~al.(2021)Zhang, Benz, Lin, Karjauv, Wu, and Kweon]{zhang2021survey}
Chaoning Zhang, Philipp Benz, Chenguo Lin, Adil Karjauv, Jing Wu, and In~So Kweon.
\newblock A survey on universal adversarial attack.
\newblock \emph{arXiv preprint arXiv:2103.01498}, 2021.

\bibitem[Zhang et~al.(2023)Zhang, Wang, Wu, and Jiang]{zhang2023diffusionad}
Hui Zhang, Zheng Wang, Zuxuan Wu, and Yu-Gang Jiang.
\newblock Diffusionad: Norm-guided one-step denoising diffusion for anomaly detection.
\newblock \emph{arXiv preprint arXiv:2303.08730}, 2023.

\bibitem[Zhang and Dai(2024)]{zhang2024gan}
Lin Zhang and Yang Dai.
\newblock A gan anomaly detection method based on multi-scale endogenous enhancement.
\newblock In \emph{Blockchain and Web3 Technology Innovation and Application Exchange Conference}, pages 269--281. Springer, 2024.

\bibitem[Zou et~al.(2022)Zou, Jeong, Pemula, Zhang, and Dabeer]{zou2022spot}
Yang Zou, Jongheon Jeong, Latha Pemula, Dongqing Zhang, and Onkar Dabeer.
\newblock Spot-the-difference self-supervised pre-training for anomaly detection and segmentation.
\newblock In \emph{European Conference on Computer Vision}, pages 392--408. Springer, 2022.

\end{thebibliography}
}

\clearpage
\setcounter{page}{1}
\maketitlesupplementary
\appendix

\section{Additional Related Work} \label{appendix:additional related work}

In this section, we shed more light on some of the previous pioneering works in AD and AL, and their pipeline.

\textbf{Recontrast} \cite{guo2024recontrast} innovates anomaly detection through contrastive reconstruction by adapting encoder and decoder networks specifically to the target domain. Unlike traditional approaches relying on frozen pre-trained encoders, it embeds contrastive learning elements into feature reconstruction to stabilize training, avoid pattern collapse, and improve domain relevance. This ensures precise anomaly detection in industrial and medical imaging tasks.

\textbf{Transformaly} \cite{cohen2021transformaly} focuses on anomaly detection using a dual-feature approach. It leverages a pre-trained ViT to extract agnostic feature vectors and employs teacher-student training to fine-tune a student network on normal samples. This complementary representation enhances anomaly detection, achieving high AUROC results in unimodal and multimodal settings.

\textbf{GeneralAD} \cite{strater2024generalad} utilizes a Vision Transformer-based framework for anomaly detection across diverse domains. It introduces a self-supervised anomaly feature generation module to create pseudo-abnormal samples by applying operations like noise addition and patch shuffling. These are fed into an attention-based discriminator to detect and localize anomalies while producing interpretable anomaly maps.

\textbf{GLASS} \cite{chen2024unified} uses gradient ascent for anomaly synthesis. This unified approach combines global anomaly synthesis to manipulate feature manifolds and local strategies to refine weak anomalies. Together, it improves the precision and breadth of industrial anomaly detection and localization.

\section{Augmentation Details} \label{appendix:augmentation}

In this section, we clarify the soft and hard augmentations, $t_i^s, t_i^h$, which are used in the foreground estimation part of our method. Here, we explain in more detail what these augmentations are and what are the rationales behind choosing them.

\textbf{Soft Augmentations. } Soft augmentations refer to transformations that do not alter the semantic content of the image, preserving the original context and interpretability of the visual information. Examples include color jitter (which modifies brightness, contrast, saturation, or hue slightly), color tint (adding a consistent color overlay), grayscale conversion (removing color information but maintaining structure), and minor Gaussian noise (introducing slight variations that mimic sensor noise). These transformations ensure that the augmented images remain perceptually similar to the originals, focusing on maintaining semantic integrity while introducing subtle variability. Such augmentations are critical in our method for refining the estimation of the foreground without distorting the regions of interest.

\textbf{Hard Augmentations. } Hard augmentations, in contrast, involve transformations that can significantly alter the semantic meaning or structure of the image. Examples include large rotations (which may distort spatial relationships), extreme cropping (removing substantial portions of the image, potentially excluding key objects), elastic transformations (which deform image structures in ways that can obscure original semantics), and heavy noise injection. These transformations challenge the robustness of the foreground estimation by introducing substantial changes, effectively creating conditions where the boundaries of semantic preservation are tested. In our method, hard augmentations are designed to evaluate the resilience of the anomaly generation process and its ability to adapt under challenging conditions.

\section{Additional Ablation Studies} \label{appendix:additional ablation}

\subsection{Clean Training} \label{appendix:pretrain ablation}
In this section, we chose to omit adversarial training and instead trained our method using standard training while keeping all other components unchanged. The results reveal an improvement in clean performance, highlighting PatchGuard's effectiveness across various training and evaluation scenarios. These results are detailed in Table \ref{tab:clean training table}. Additionally, we evaluated the clean-trained model under an adversarial setup, demonstrating that our pipeline benefits significantly from adversarial training. This underscores the impact of our regularization technique in adversarial training, which enhances the robustness of attention-based mechanisms against adversarial examples.

\begin{table}[h]
    \caption{Performance of the model trained without adversarial training under clean and adversarial setups.}
    
    \resizebox{ \columnwidth}{!}
    {\begin{tabular}{@{}cccccc} 
    
    \specialrule{1.5pt}{\aboverulesep}{\belowrulesep}
    \multirow{2}{*}{\textbf{Method}} & \multirow{2}{*}{\textbf{Task}} & \multicolumn{4}{c}{\textbf{Dataset}} \\
      \cmidrule(lr){3-6}

\textbf{} && \textbf{MVTec AD}& \textbf{VisA}& \textbf{BTAD} & \textbf{BraTS2021}\\ 

    \specialrule{1.5pt}{\aboverulesep}{\belowrulesep} 

    \noalign{\vskip 1.5pt}

   \multirow{2}{*}{Clean} 
   & AD & \textcolor{gray}{94.7 / }12.9 & \textcolor{gray}{93.9 / }16.3 & \textcolor{gray}{91.3 / }11.6 & \textcolor{gray}{97.1 / }10.7\\

   \cmidrule(lr){3-6}

    & AL & \textcolor{gray}{97.4 / }12.5 & \textcolor{gray}{98.0 / }8.1 & \textcolor{gray}{95.7 / }11.6 & \textcolor{gray}{98.5 / }12.3\\

   \noalign{\vskip 1.5pt} 
    \specialrule{1.5pt}{\aboverulesep}{\belowrulesep}
     \end{tabular}}
    \label{tab:clean training table}

\end{table}

\subsection{Ablation on $\delta$} \label{appendix:vit ablation}
To determine the attention degree for an output token in an attention head, you need to identify how many of the total input tokens it attends to more than the others. In our case, the ViT model has 256 input tokens. Intuitively, we set \( \delta \) to \( \frac{1}{255} \), drawn from a uniform distribution. In Table \ref{tab:delta ablation}, we present an ablation study on the value of \( \delta \).

\begin{table}[h]
    \caption{Ablation study on the value of \( \delta \) and its effect on the attention degree in the ViT model.}
    
    \resizebox{ \columnwidth}{!}
    {\begin{tabular}{@{}cccccc} 
    
    \specialrule{1.5pt}{\aboverulesep}{\belowrulesep}
    \multirow{2}{*}{\textbf{$\delta$}} & \multirow{2}{*}{\textbf{Task}} & \multicolumn{4}{c}{\textbf{Dataset}} \\
      \cmidrule(lr){3-6}

\textbf{} && \textbf{MVTec AD}& \textbf{VisA}& \textbf{BTAD} & \textbf{BraTS2021}\\ 

    \specialrule{1.5pt}{\aboverulesep}{\belowrulesep} 

    \noalign{\vskip 1.5pt}

   \multirow{2}{*}{$\frac{1}{2 \times 256}$} & AD & \textcolor{gray}{89.0 / }69.1 & \textcolor{gray}{88.8 / }72.1 & \textcolor{gray}{86.7 / }79.4 & \textcolor{gray}{94.4 / }79.4 \\

   \cmidrule(lr){3-6}

   & AL & \textcolor{gray}{93.1 / }70.9 & \textcolor{gray}{96.8 / }82.0 & \textcolor{gray}{94.0 / }70.8 & \textcolor{gray}{97.8 / }90.1\\

    \noalign{\vskip 1.5pt} 
    \specialrule{1pt}{\aboverulesep}{\belowrulesep}
    \noalign{\vskip 1.5pt} 

   \multirow{2}{*}{$\frac{1}{256}$} & AD & \textcolor{gray}{88.1 / }71.1 & \textcolor{gray}{88.5 / }74.3 & \textcolor{gray}{85.3 / }82.1 & \textcolor{gray}{94.3 / }81.0\\

   \cmidrule(lr){3-6}

   & AL & \textcolor{gray}{92.7 / }73.8 & \textcolor{gray}{96.9 /}85.2 & \textcolor{gray}{93.2  /}73.0 & \textcolor{gray}{97.7 /}94.5\\

   \noalign{\vskip 1.5pt} 
    \specialrule{1pt}{\aboverulesep}{\belowrulesep}
    \noalign{\vskip 1.5pt} 

   \multirow{2}{*}{$\frac{2}{256}$} & AD & \textcolor{gray}{85.1 / }71.9 & \textcolor{gray}{84.7 / }75.8 & \textcolor{gray}{82.4 / }82.8 & \textcolor{gray}{91.7 / }81.3 \\

   \cmidrule(lr){3-6}

   & AL & \textcolor{gray}{88.6 / }74.3 & \textcolor{gray}{93.4 / }85.6 & \textcolor{gray}{90.3 / }74.1 & \textcolor{gray}{92.7 / }81.9\\

   \noalign{\vskip 1.5pt} 
    \specialrule{1.5pt}{\aboverulesep}{\belowrulesep}
     \end{tabular}}
    \label{tab:delta ablation}

\end{table}

\subsection{Diffenet ViT Backbone} \label{appendix:pretrain ablation}
As mentioned in the implementation details, we used a ViT small model with a patch size of 14, initialized with random weights. In this section, we evaluate our method by replacing the backbone with larger variations of ViT models (note that these models use random weights and are not pre-trained). All other components are kept fixed. As shown in Table \ref{tab:pretrain table}, the results demonstrate that our method achieves high performance and consistency across different backbones.

\begin{table}[h]
    \caption{Evaluation of our method with different ViT backbones initialized with random weights. The results demonstrate high performance and consistency across various backbone configurations.}
    
    \resizebox{ \columnwidth}{!}
    {\begin{tabular}{@{}cccccc} 
    
    \specialrule{1.5pt}{\aboverulesep}{\belowrulesep}
    \multirow{2}{*}{\textbf{ViT}} & \multirow{2}{*}{\textbf{Task}} & \multicolumn{4}{c}{\textbf{Dataset}} \\
      \cmidrule(lr){3-6}

\textbf{} && \textbf{MVTec AD}& \textbf{VisA}& \textbf{BTAD} & \textbf{BraTS2021}\\ 

    \specialrule{1.5pt}{\aboverulesep}{\belowrulesep} 

    \noalign{\vskip 1.5pt}

    \multirow{2}{*}{Small(\textit{Ours})} & AD & \textcolor{gray}{88.1 / }71.1 & \textcolor{gray}{88.5 / }74.3 & \textcolor{gray}{85.3 / }82.1 & \textcolor{gray}{94.3 / }81.0\\

   \cmidrule(lr){3-6}

   & AL & \textcolor{gray}{92.7 / }73.8 & \textcolor{gray}{96.9 /}85.2 & \textcolor{gray}{93.2  /}73.0 & \textcolor{gray}{97.7 /}94.5\\

    \noalign{\vskip 1.5pt} 
    \specialrule{1pt}{\aboverulesep}{\belowrulesep}
    \noalign{\vskip 1.5pt} 

   \multirow{2}{*}{Base} & AD & \textcolor{gray}{89.1 / }71.0 & \textcolor{gray}{87.9 / }73.1 & \textcolor{gray}{84.5 / }81.7 & \textcolor{gray}{93.0 / }82.3 \\

   \cmidrule(lr){3-6}

   & AL & \textcolor{gray}{91.0 / }72.6 & \textcolor{gray}{95.8 / }82.1 & \textcolor{gray}{94.7 / }74.8 & \textcolor{gray}{98.4 / }94.9 \\

   \noalign{\vskip 1.5pt} 
    \specialrule{1pt}{\aboverulesep}{\belowrulesep}
    \noalign{\vskip 1.5pt} 

   \multirow{2}{*}{Large} & AD & \textcolor{gray}{90.0 / }70.6 & \textcolor{gray}{87.5 / }74.7 & \textcolor{gray}{85.5 / }81.9 & \textcolor{gray}{95.1 / }81.6 \\

   \cmidrule(lr){3-6}

   & AL & \textcolor{gray}{92.9 / }73.4 & \textcolor{gray}{95.8 / }86.0 & \textcolor{gray}{94.1 / }73.5 & \textcolor{gray}{96.7 / }93.2\\

   \noalign{\vskip 1.5pt} 
    \specialrule{1.5pt}{\aboverulesep}{\belowrulesep}
     \end{tabular}}
    \label{tab:pretrain table}

\end{table}

\subsection{Backbone} \label{appendix:pretrain ablation}
\begin{table}[b]
    \caption{A study on the performance of various backbone networks, as alternatives to the ViT, within our architecture.}
     
     \resizebox{ \columnwidth}{!}{\begin{tabular}{@{}cccccc@{}} 

    \specialrule{1.5pt}{\aboverulesep}{\belowrulesep}
    \multirow{2}{*}{\textbf{Backbone}} & \multirow{2}{*}{\textbf{Task}} & \multicolumn{4}{c}{\textbf{Dataset}}
    \\  \cmidrule(lr{0pt}){3-6} 

    &&  \textbf{MVTec-AD} & \textbf{VisA} & \textbf{BTAD} & \textbf{BraTS2021}  \\

    \specialrule{1.5pt}{\aboverulesep}{\belowrulesep}

    \noalign{\vskip 3pt} 
    
    \multirow{2}{*}{\textbf{U-Net}} & AD & \graytext{84.7 /} 15.1 & \graytext{80.0 /} 15.7 & \graytext{76.9 /} 14.0 & \graytext{70.3 /} 12.9 \\

    \cmidrule(lr{0pt}){3-6}

    & AL & \graytext{85.4 /} 17.8 & \graytext{81.8 /} 13.9 & \graytext{79.6 /} 18.2 & \graytext{76.3 /} 16.0 \\

    \noalign{\vskip 1.5pt} 
    \specialrule{1pt}{\aboverulesep}{\belowrulesep}
    \noalign{\vskip 1.5pt} 

    \multirow{2}{*}{\textbf{Resnet50}} & AD & \graytext{88.1 /} 27.6 & \graytext{84.9 /} 23.3 & \graytext{85.7 /} 23.9 & \graytext{86.0 /} 23.5 \\

    \cmidrule(lr{0pt}){3-6}

    & AL & \graytext{87.1 /} 28.3 & \graytext{83.7 /} 24.6 & \graytext{86.0 /} 24.7 & \graytext{85.1 /} 23.9 \\

    \noalign{\vskip 3pt}

    \specialrule{1.5pt}{\aboverulesep}{\belowrulesep}

     \end{tabular}}
    \label{tab:ablation backbone}

\end{table}

In Sections \ref{theory} and \ref{method}, we provided intuitions and theoretical insights on why vision transformers achieve better adversarial robustness than convolution-based methods. In this section, we use convolution-based backbones in our pipeline instead on the ViT, while preserving all other components as they are. To support this claim, we provide the detection and localization results in Table \ref{tab:ablation backbone}.

Adapting convolution-based backbones like ResNet \cite{resnet} to our patch-based pipeline poses certain challenges. To integrate ResNet, we incorporate a binary classification layer at the model's final stage. Each image is divided into patches manually, following the same patching approach used by the vision transformer. Anomaly scores are then computed for each patch independently, and the final anomaly detection decision is based on the top-$k$ patches with the highest scores. For a fair comparison, we apply the same hyperparameters used in our original method.

To adapt U-Net \cite{ronneberger2015u}, we shift from a patch-wise approach to pixel-wise localization, given the architectural constraints of U-Net. Notably, the top-$k$ selection used previously is incompatible in this context. Instead, we employ a top-$p$ percent pixel selection for the anomaly detection decision, where $p = \frac{k}{N},$ and
$N$ represents the total number of patches in an image.

\subsection{Integrating Sparse Attention Mechanism into Our Methodology} \label{appendix:Applying-Sparse-Attention}
We evaluate the performance of our method after incorporating BigBird \cite{zaheer2021bigbirdtransformerslonger}, a sparse attention mechanism, as shown in Table \ref{tab:sparse-attention}. The results reveal two key findings. First, applying the sparse attention mechanism generally reduces the robustness of our method. This aligns with our intuition and theory, which suggest that a higher attention degree enhances model robustness, while sparse attention decreases the attention degree. Second, our regularization term remains effective even in the sparse attention setup—when applied, it still improves the model's robustness.

\begin{table}[th]
    \caption{Performance comparison of our method with and without the BigBird sparse attention mechanism.}
    \label{tab:sparse-attention}
    \resizebox{\columnwidth}{!}{%
    \begin{tabular}{@{}cccccc@{}}
        \specialrule{0.5pt}{\aboverulesep}{\belowrulesep}
        \textbf{Backbone} & \textbf{Task} & \textbf{MVTec AD} & \textbf{VisA} & \textbf{BTAD} & \textbf{BraTS2021} \\
        \specialrule{0.5pt}{\aboverulesep}{\belowrulesep}
        BigBird & AD & \graytext{86.7} / 51.7 & \graytext{90.3} / 49.7 & \graytext{86.0} / 58.9 & \graytext{95.9} / 61.8 \\
        & AL & \graytext{91.4} / 53.1 & \graytext{93.9} / 59.8 & \graytext{92.5} / 51.4 & \graytext{96.4} / 68.5 \\
        \specialrule{0.5pt}{\aboverulesep}{\belowrulesep}
        BigBird + Our Regularization & AD & \graytext{85.6} / 63.0 & \graytext{88.5} / 61.4 & \graytext{86.2} / 69.8 & \graytext{92.2} / 73.1 \\
         & AL & \graytext{90.0} / 64.7 & \graytext{92.1} / 73.6 & \graytext{90.5} / 62.2 & \graytext{94.6} / 78.9 \\
        \specialrule{0.5pt}{\aboverulesep}{\belowrulesep}
        Ours & AD & \graytext{88.1} / 71.1 & \graytext{88.5} / 74.3 & \graytext{85.3} / 82.1 & \graytext{94.3} / 81.0 \\
             & AL & \graytext{92.7} / 73.8 & \graytext{96.9} / 85.2 & \graytext{93.2} / 73.0 & \graytext{97.7} / 94.5 \\
        \specialrule{0.5pt}{\aboverulesep}{\belowrulesep}
    \end{tabular}
    }
\end{table}

\subsection{Impact of Regularization Layer on Model Performance}
We investigated the effect of applying regularization to different layers of the network. The results, shown in Table \ref{tab:reg-layer-impact}, indicate that regularization in inner layers generally improves model robustness. However, the last layer performs slightly better according to our experiments.

\begin{table}[h]
    \caption{Performance of Regularization at Different Layers}
    \label{tab:reg-layer-impact}
    \centering
    \resizebox{\columnwidth}{!}{%
    \begin{tabular}{@{}cccccc@{}}
        \specialrule{0.5pt}{\aboverulesep}{\belowrulesep}
        \textbf{Regularization Layer} & \textbf{Task} & \textbf{MVTec AD} & \textbf{VisA} & \textbf{BTAD} & \textbf{BraTS2021} \\
        \specialrule{0.5pt}{\aboverulesep}{\belowrulesep}
        $(N-2)^{\text{th}}$ & AD & \graytext{87.3} / 68.2 & \graytext{89.7} / 72.9 & \graytext{86.3} / 79.1 & \graytext{93.5} / 75.4 \\
        & AL & \graytext{91.5} / 70.6 & \graytext{96.0} / 83.1 & \graytext{91.7} / 68.5 & \graytext{96.8} / 90.6 \\
        \specialrule{0.5pt}{\aboverulesep}{\belowrulesep}
        $(N-1)^{\text{th}}$ & AD & \graytext{89.0} / 69.4 & \graytext{87.5} / 73.1 & \graytext{84.3} / 82.3 & \graytext{92.6} / 79.9 \\
        & AL & \graytext{93.2} / 70.6 & \graytext{95.7} / 86.1 & \graytext{93.1} / 71.7 & \graytext{97.0} / 93.2 \\
        \specialrule{0.5pt}{\aboverulesep}{\belowrulesep}
        $N^{\text{th}}$ (Last Layer) & AD & \graytext{88.1} / 71.1 & \graytext{88.5} / 74.3 & \graytext{85.3} / 82.1 & \graytext{94.3} / 81.0 \\
         & AL & \graytext{92.7} / 73.8 & \graytext{96.9} / 85.2 & \graytext{93.2} / 73.0 & \graytext{97.7} / 94.5 \\
        \specialrule{0.5pt}{\aboverulesep}{\belowrulesep}
    \end{tabular}
    }
\end{table}

\section{Dataset Details} \label{appendix:dataset details}
We conducted our experiments on eight datasets covering a diverse range of domains, from industrial to medical applications. The medical datasets include BraTS2021 and Head-CT, while the remaining six datasets focus on industrial and synthetic anomaly detection and localization tasks. Below, we provide detailed descriptions of each dataset.

\begin{itemize}
\item MVTec AD:
MVTec AD is a dataset for benchmarking anomaly detection methods in industrial inspection. It includes over 5,000 high-resolution images across fifteen object and texture categories. Each category contains defect-free training images and a test set with both normal and defective samples, featuring defects like scratches, dents, and misalignments.

\item VisA:
The VisA dataset contains 12 object subsets with 10,821 images, comprising 9,621 normal samples and 1,200 anomalous samples. The subsets include printed circuit boards, multi-instance objects like Capsules and Macaroni, and roughly aligned objects such as Cashew and Chewing Gum. Anomalies include surface defects like scratches and dents, and structural issues such as missing components.

\item BTAD:
The BTAD \cite{mishra21-vt-adl} consists of 2,830 images of three industrial products. It provides samples with body and surface defects, intended for evaluating visual anomaly detection methods in industrial settings.

\item MPDD:
MPDD \cite{9631567} is a dataset for defect detection in metal parts manufacturing, consisting of over 1,000 images with pixel-level defect annotations. The dataset is divided into six distinct classes and includes anomaly-free training samples and test samples with normal and defective parts, covering a variety of surface and structural defects.

\item WFDD:
WFDD \cite{chen2024unified} is a dataset for anomaly detection in textile inspection, comprising 4,101 woven fabric images across four categories: grey cloth, grid cloth, yellow cloth, and pink flower. Defects are categorized as block-shaped, point-like, or line-type, with pixel-level annotations.

\item DTD-Synthetic:
The DTD-Synthetic \cite{Aota_2023_WACV} is designed for anomaly detection and segmentation tasks, containing synthetic texture images generated from predefined texture patterns. It includes twelve classes of normal texture samples and those with artificially introduced anomalies such as structural distortions or irregular patterns.

\item BraTS2021:
BraTS2021 is a medical dataset for anomaly segmentation, containing 1,251 MRI cases with voxel-level annotations for tumor regions. Each case includes multiple imaging modalities (T1, T1ce, T2, and FLAIR). In this paper, only the FLAIR modality is used due to its sensitivity to tumor regions.

\item Head-CT:
The Head-CT \cite{kitamura2018headct} contains 200 head CT slices, evenly split between normal slices and those with hemorrhages, without distinguishing between hemorrhage types.

\end{itemize}

\section{Details of Adaptation of State-of-the-Art Methods for Adversarial Robustness} \label{appendix:detail-adaptation-sota-method-to-adv-training} 

Before proposing a novel approach for adversarial AD and AL setups, our idea was to adapt existing state-of-the-art methods in the field and enhance their robustness through adversarial training \cite{strater2024generalad, chen2024unified, yao2024glad, guo2024recontrast}. Adversarial training involves feeding adversarial examples to the model during training \cite{madry2017towards}. In the following section, we explain how we create adversarial examples for each method and the details of our best approach to make them robust, which are reported in Table \ref{tab:adapting clean methods table}.

Anomaly-free methods like PatchCore \cite{roth2021total} and ReContrast \cite{guo2024recontrast} have anomaly-free training. For adapting PatchCore, we used an adversarially trained ResNet-50 \cite{resnet} as the feature extractor. Adding adversarially generated normal samples to the memory bank was tested but did not improve performance. For adapting ReContrast, we replaced both the teacher and student networks with adversarially trained ResNet-50 models, and using adversarial samples generated by the PGD-100 attack \cite{madry2017towards} on the final anomaly map, we trained the network to improve robustness.

Embedding-space synthesis methods, such as GeneralAD \cite{strater2024generalad} and SimpleNet \cite{liu2023simplenet}, operate in the embedding space. For adapting SimpleNet, we employed an adversarially robust WideResNet-50 \cite{zagoruyko2017wide} as the feature extractor. Two strategies were tested: input-space adversarial training, which was ineffective due to the absence of anomaly samples, and embedding-space adversarial training, where the discriminator was adversarially trained using both normal and anomaly features. The latter approach performed better, as reported in Table \ref{tab:adapting clean methods table}, but the overall performance was still insufficient. For GeneralAD, due to the model's dependency on DINO \cite{caron2020unsupervised} pre-trained weights, we could not find a proper robust pre-trained ViT backbone for adaptation that maintained clean performance. We tested multiple approaches to make it robust while maintaining clean performance, and the best one was embedding-space adversarial training of the discriminator with access to both normal and anomaly features, and replacing the ViT with a robust pre-trained model on ImageNet \cite{deng2009imagenet}. 

Input-space synthesis methods like DRAEM \cite{zavrtanik2021draem} and GLASS \cite{chen2024unified} generate synthetic anomalies in the input space. In adapting DRAEM, adversarial samples were created using PGD-100 on the focal loss \cite{lin2017focal} of the anomaly map, and these samples were used to train both the reconstructive and discriminative sub-networks adversarially. For adapting GLASS, the best results were achieved by combining an adversarially trained feature extractor with adversarial samples (PGD-100) applied to both \( L_n \) and \( L_{las} \).

According to Table \ref{tab:adapting clean methods table}, state-of-the-art methods in AD and AL, even after adapting to adversarial training scenarios, still suffer from vulnerability to adversarial attacks and perform weakly.

\section{Per-Class Results} \label{appendix:per-class results}

In this section, we present the per-class AUROC results for anomaly detection and localization using PatchGuard across the reported datasets, as detailed in Tables~\ref{tab:class_wise_mvtec}, \ref{tab:class_wise_visa}, \ref{tab:class_wise_btad}, \ref{tab:class_wise_mpdd}, \ref{tab:class_wise_wfdd}, and \ref{tab:class_wise_dtd}.

\begin{table}[t]
    \caption{Class-wise Clean and Adversarial AUROC (\%) Results for Image-level and Pixel-level Evaluations on the MVTec-AD Dataset.}
    
    \resizebox{\columnwidth}{!}
    {\begin{tabular}{@{}cccccc@{}} 
    
    \specialrule{1.5pt}{\aboverulesep}{\belowrulesep}
    \textbf{Class Name} & \multicolumn{2}{c}{\textbf{Image-level AUROC (\%)}} & \multicolumn{2}{c}{\textbf{Pixel-level AUROC (\%)}} \\
    \cmidrule(lr){2-3} \cmidrule(lr){4-5}
    & \textbf{Clean} & \textbf{Adversarial} & \textbf{Clean} & \textbf{Adversarial} \\
    
    \specialrule{1.5pt}{\aboverulesep}{\belowrulesep} 
    
    Bottle & 97.6 & 84.7 & 96.7 & 84.6 \\ 
    Cable & 87.3 & 74.0 & 97.2 & 77.8 \\ 
    Capsule & 71.8 & 79.6 & 93.6 & 85.2 \\ 
    Carpet & 83.9 & 43.2 & 95.8 & 53.7 \\ 
    Grid & 95.7 & 74.9 & 93.1 & 55.5 \\ 
    Hazelnut & 99.5 & 80.5 & 97.2 & 91.3 \\ 
    Leather & 91.0 & 80.2 & 97.6 & 67.6 \\ 
    Metal Nut & 88.8 & 54.4 & 87.9 & 75.3 \\ 
    Pill & 81.5 & 59.1 & 86.7 & 76.3 \\ 
    Screw & 55.4 & 56.6 & 93.8 & 86.5 \\ 
    Tile & 95.3 & 71.7 & 86.2 & 64.0 \\ 
    Toothbrush & 100 & 90.6 & 93.9 & 81.9 \\ 
    Transistor & 94.1 & 84.3 & 95.4 & 84.8 \\ 
    Wood & 93.1 & 56.0 & 89.6 & 57.4 \\ 
    Zipper & 87.6 & 76.9 & 86.2 & 65.7 \\

    \specialrule{1.5pt}{\aboverulesep}{\belowrulesep}
    
    \end{tabular}}
    \label{tab:class_wise_mvtec}
\end{table}

\begin{table}[t]
    \caption{Class-wise Clean and Adversarial AUROC (\%) Results for Image-level and Pixel-level Evaluations on the VisA Dataset.}
    
    \resizebox{\columnwidth}{!}
    {\begin{tabular}{@{}cccccc@{}} 
    
    \specialrule{1.5pt}{\aboverulesep}{\belowrulesep}
    \textbf{Class Name} & \multicolumn{2}{c}{\textbf{Image-level AUROC (\%)}} & \multicolumn{2}{c}{\textbf{Pixel-level AUROC (\%)}} \\
    \cmidrule(lr){2-3} \cmidrule(lr){4-5}
    & \textbf{Clean} & \textbf{Adversarial} & \textbf{Clean} & \textbf{Adversarial} \\
    
    \specialrule{1.5pt}{\aboverulesep}{\belowrulesep} 
    
    Candle & 83.6 & 82.5 & 94.9 & 70.7 \\ 
    Capsules & 77.7 & 66.2 & 97.1 & 57.0 \\ 
    Cashew & 88.7 & 83 & 97.3 & 89.9 \\ 
    Chewing gum & 92.2 & 73.5 & 97.7 & 92.2 \\ 
    Fryum & 85.1 & 74.5 & 95.9 & 88.0 \\ 
    Macaroni 1 & 86.5 & 66.2 & 97.0 & 84.8 \\ 
    Macaroni 2 & 68.3 & 42.3 & 95.3 & 85.0 \\ 
    Pcb 1 & 95.4 & 85.3 & 99.0 & 95.4 \\ 
    Pcb 2 & 97.3 & 91.2 & 96.8 & 87.1 \\ 
    Pcb 3 & 94.8 & 73.5 & 98.7 & 93.0 \\ 
    Pcb 4 & 98.5 & 92.1 & 95.9 & 83.7 \\ 
    Pipe fryum & 94.3 & 61.5 & 98.1 & 96.2 \\

    \specialrule{1.5pt}{\aboverulesep}{\belowrulesep}
    
    \end{tabular}}
    \label{tab:class_wise_visa}
\end{table}

\begin{table}[t]
    \caption{Class-wise Clean and Adversarial AUROC (\%) Results for Image-level and Pixel-level Evaluations on the BTAD Dataset.}
    
    \resizebox{\columnwidth}{!}
    {\begin{tabular}{@{}cccccc@{}} 
    
    \specialrule{1.5pt}{\aboverulesep}{\belowrulesep}
    \textbf{Class Name} & \multicolumn{2}{c}{\textbf{Image-level AUROC (\%)}} & \multicolumn{2}{c}{\textbf{Pixel-level AUROC (\%)}} \\
    \cmidrule(lr){2-3} \cmidrule(lr){4-5}
    & \textbf{Clean} & \textbf{Adversarial} & \textbf{Clean} & \textbf{Adversarial} \\
    
    \specialrule{1.5pt}{\aboverulesep}{\belowrulesep} 
    
    01 & 98.6 & 96.1 & 91.7 & 77.1 \\ 
    02 & 65.3 & 65.6 & 92.1 & 64.2 \\ 
    03 & 92.0 & 84.6 & 95.8 & 77.8 \\

    \specialrule{1.5pt}{\aboverulesep}{\belowrulesep}
    
    \end{tabular}}
    \label{tab:class_wise_btad}
\end{table}

\begin{table}[t]
    \caption{Class-wise Clean and Adversarial AUROC (\%) Results for Image-level and Pixel-level Evaluations on the MPDD Dataset.}
    
    \resizebox{\columnwidth}{!}
    {\begin{tabular}{@{}cccccc@{}} 
    
    \specialrule{1.5pt}{\aboverulesep}{\belowrulesep}
    \textbf{Class Name} & \multicolumn{2}{c}{\textbf{Image-level AUROC (\%)}} & \multicolumn{2}{c}{\textbf{Pixel-level AUROC (\%)}} \\
    \cmidrule(lr){2-3} \cmidrule(lr){4-5}
    & \textbf{Clean} & \textbf{Adversarial} & \textbf{Clean} & \textbf{Adversarial} \\
    
    \specialrule{1.5pt}{\aboverulesep}{\belowrulesep} 
    
    Bracket Black & 83.4 & 60.1 & 92.1 & 88.4 \\ 
    Bracket Brown & 84.5 & 77.4 & 91.0 & 84.7 \\ 
    Bracket White & 79.8 & 52.9 & 92.1 & 85.9 \\ 
    Connector & 94.3 & 92.5 & 93.8 & 76.8 \\ 
    Metal Plate & 100 & 86.2 & 98.2 & 95.2 \\ 
    Tubes & 70.4 & 42.8 & 95.9 & 88.9 \\

    \specialrule{1.5pt}{\aboverulesep}{\belowrulesep}
    
    \end{tabular}}
    \label{tab:class_wise_mpdd}
\end{table}

\begin{table}[t]
    \caption{Class-wise Clean and Adversarial AUROC (\%) Results for Image-level and Pixel-level Evaluations on the WFDD Dataset.}
    
    \resizebox{\columnwidth}{!}
    {\begin{tabular}{@{}cccccc@{}} 
    
    \specialrule{1.5pt}{\aboverulesep}{\belowrulesep}
    \textbf{Class Name} & \multicolumn{2}{c}{\textbf{Image-level AUROC (\%)}} & \multicolumn{2}{c}{\textbf{Pixel-level AUROC (\%)}} \\
    \cmidrule(lr){2-3} \cmidrule(lr){4-5}
    & \textbf{Clean} & \textbf{Adversarial} & \textbf{Clean} & \textbf{Adversarial} \\
    
    \specialrule{1.5pt}{\aboverulesep}{\belowrulesep} 
    
    Gray Cloth & 88.8 & 57.3 & 93.1 & 75.9 \\ 
    Grid Cloth & 99.6 & 94.7 & 97.8 & 85.1 \\ 
    Pink Flower & 52.3 & 33.0 & 94.4 & 63.0 \\ 
    Yellow Cloth & 96.3 & 75.0 & 93.3 & 62.5 \\

    \specialrule{1.5pt}{\aboverulesep}{\belowrulesep}
    
    \end{tabular}}
    \label{tab:class_wise_wfdd}
\end{table}

\begin{table}[t]
    \caption{Class-wise Clean and Adversarial AUROC (\%) Results for Image-level and Pixel-level Evaluations on the DTD-Synthetic Dataset.}
    
    \resizebox{\columnwidth}{!}
    {\begin{tabular}{@{}cccccc@{}} 
    
    \specialrule{1.5pt}{\aboverulesep}{\belowrulesep}
    \textbf{Class Name} & \multicolumn{2}{c}{\textbf{Image-level AUROC (\%)}} & \multicolumn{2}{c}{\textbf{Pixel-level AUROC (\%)}} \\
    \cmidrule(lr){2-3} \cmidrule(lr){4-5}
    & \textbf{Clean} & \textbf{Adversarial} & \textbf{Clean} & \textbf{Adversarial} \\
    
    \specialrule{1.5pt}{\aboverulesep}{\belowrulesep} 
    
    Blotchy 099 & 86.1 & 73.2 & 94.7 & 87.5 \\ 
    Fibrous 183 & 100 & 55.8 & 99.0 & 80.9 \\ 
    Marbled 078 & 89.3 & 63.8 & 97.6 & 88.4 \\ 
    Matted 069 & 93.1 & 53.8 & 97.1 & 78.1 \\ 
    Mesh 114 & 94.4 & 74.9 & 97.6 & 73.6 \\ 
    Perforated 037 & 99.9 & 81.1 & 94.4 & 55.5 \\ 
    Stratified 154 & 91.6 & 42.5 & 98.2 & 76.2 \\ 
    Woven 001 & 83.6 & 56.0 & 96.2 & 71.6 \\ 
    Woven 068 & 88.5 & 55.5 & 93.6 & 71.4 \\ 
    Woven 104 & 79.6 & 55.4 & 87.9 & 73.0 \\ 
    Woven 125 & 95.3 & 56.9 & 98.2 & 71.6 \\ 
    Woven 127 & 96.8 & 57.9 & 97.2 & 70.5 \\

    \specialrule{1.5pt}{\aboverulesep}{\belowrulesep}
    
    \end{tabular}}
    \label{tab:class_wise_dtd}
\end{table}

\section{Implementation Details}
\label{appendix:implementation}

The optimizer used is \texttt{AdamW} \cite{loshchilov2018decoupled}, with a learning rate of 0.0008 and a weight decay of 0.00001. For learning rate scheduling, we utilize a \texttt{CosineAnnealingLR} scheduler with a decay factor of 0.0125, where the minimum learning rate ($\eta_{\text{min}}$) is calculated as \texttt{lr} $\times$ \texttt{lr\_decay\_factor}, and \texttt{T\_max} is set to the number of epochs, which is set to 300 but we observe empirically that convergence usually happens much faster. The batch size for both training and testing is set to 16. The input image size is $224 \times 224$. We use a \texttt{ViT} (Vision Transformer) small model as the feature extractor, which is not pre-trained. Finally, we perform a top-$k$ selection of the localization map achieved, to obtain a final AD decision, with $k$ being set to 5.

\section{Attack Adaptation Details} \label{appendix:attack adaptation}
\textbf{Adaptation Classification Attack.} We evaluated PatchGuard's resilience against several advanced adapted attacks from the classification domain, including CAA, AutoAttack, A3, and PGD-1000. Originally designed to compromise classification tasks by exploiting the cross-entropy loss, these attacks were adapted for anomaly localization (AL) and anomaly detection (AD) tasks. The focus was on altering the sum of cross-entropy for all patches in detector models, aiming to increase the loss values for normal regions of test samples while decreasing them for anomalous regions. Adapting AutoAttack (AA) for AD and AL tasks posed significant challenges. AutoAttack comprises a suite of different attack methods, such as FAB, multi-targeted FAB, Square Attack, APGDT, APGD with cross-entropy loss, and APGD with DLR loss. The primary difficulty in adaptation arises because attacks using the DLR loss assume the model’s output contains at least three elements, an assumption valid for classification tasks with three or more classes but not applicable to AD and AL tasks. Consequently, we replaced the DLR loss component in AutoAttack with a PGD attack. However, for the other attacks under consideration, no modifications were necessary.

\textbf{Adaptation Segmentation Attack.} We evaluate our method against advanced adapted semantic segmentation attacks, specifically SegPGD and SEA, with a key modification: instead of operating at the pixel level, our approach applies these attacks patch-wise. SegPGD is a segmentation-specific adaptation of the Projected Gradient Descent (PGD) attack that dynamically balances focus between misclassified and correctly classified pixels. It starts by prioritizing correctly classified pixels, progressively shifting its emphasis to achieve an effective balance as the attack unfolds. On the other hand, SEA integrates multiple complementary loss functions, such as Jensen-Shannon divergence and Masked Cross-Entropy, to exploit various weaknesses in model robustness. Through progressive radius reduction and adaptive optimization, SEA generates potent adversarial perturbations, selecting the worst-case attack outcome to ensure a thorough robustness evaluation.


\section{Attention Discriminator} \label{appendix:discriminator ablation}

The attention discriminator in our method plays a pivotal role in the anomaly detection and localization pipeline. Conceptually, this component operates similarly to a single layer of a Vision Transformer, where the input embeddings undergo self-attention operations. The resulting embeddings are subsequently passed through a Multi-Layer Perceptron (MLP) to compute a set of anomaly scores for each embedding. Furthermore, the ``Attention Degrees," introduced in previous sections, are derived directly from this attention discriminator, reinforcing its critical position in our framework.

To substantiate the necessity and efficacy of the attention discriminator, we performed an ablation study, the results of which are presented in Table \ref{tab:attention discriminator table}. In the alternative setup without the discriminator, the attention degrees and MLP components are instead placed on top of the ViT’s final layer. This bypass eliminates the intermediate role played by the attention discriminator. However, the comparative results demonstrate that the inclusion of the attention discriminator provides marginally superior performance. This advantage highlights its significance not only in enhancing the model’s performance but also in improving its interpretability by providing more precise and structured attention degree calculations.

\begin{table}[t]
    \caption{Comparison of our model's performance with and without the attention discriminator.}
    
    \resizebox{ \columnwidth}{!}
    {\begin{tabular}{@{}cccccc} 
    
    \specialrule{1.5pt}{\aboverulesep}{\belowrulesep}
    \multirow{2}{*}{\textbf{Method}} & \multirow{2}{*}{\textbf{Task}} & \multicolumn{4}{c}{\textbf{Dataset}} \\
      \cmidrule(lr){3-6}

&& \textbf{MVTec AD}& \textbf{VisA}& \textbf{BTAD} & \textbf{BraTS2021}\\ 

    \specialrule{1.5pt}{\aboverulesep}{\belowrulesep} 

    \noalign{\vskip 1.5pt}

   \multirow{2}{*}{w/o discriminator} & AD & \textcolor{gray}{85.9 / }69.5 & \textcolor{gray}{87.7 / }73.0 & \textcolor{gray}{83.8 / }80.6 & \textcolor{gray}{93.4 / }80.6\\

   \cmidrule(lr){3-6}

   & AL & \textcolor{gray}{91.1 / }71.5 & \textcolor{gray}{95.4 / }83.9 & \textcolor{gray}{91.7  /}72.2 & \textcolor{gray}{96.4 / }93.5\\

   \noalign{\vskip 1.5pt} 
    \specialrule{1pt}{\aboverulesep}{\belowrulesep}
    \noalign{\vskip 1.5pt}

   \multirow{2}{*}{w/ discriminator (\textit{Ours})} & AD & \textcolor{gray}{88.1 / }71.1 & \textcolor{gray}{88.5 / }74.3 & \textcolor{gray}{85.3 / }82.1 & \textcolor{gray}{94.3 / }81.0\\

   \cmidrule(lr){3-6}

   & AL & \textcolor{gray}{92.7 / }73.8 & \textcolor{gray}{96.9 / }85.2 & \textcolor{gray}{93.2  /}73.0 & \textcolor{gray}{97.7 / }94.5\\

   \noalign{\vskip 1.5pt} 
    \specialrule{1.5pt}{\aboverulesep}{\belowrulesep}
     \end{tabular}}
    \label{tab:attention discriminator table}

\end{table}

\section{Evaluating Our Model Under Various Attacks with Diverse Epsilon} \label{appendix: various attacks diverse epsilon}
To demonstrate our model's robustness, we conducted an experiment in which we trained it under varying $\epsilon$ values of PGD with $l_{\infty}$ norm and evaluated it using the same $\epsilon$ (ensuring that the training and evaluation $\epsilon$ were identical). The results, as presented in Table \ref{tab:diff_eps_eval}, indicate that PatchGuard performs effectively across different $\epsilon$ values.

\begin{table}[h]
    \caption{Performance of PatchGuard under varying $\epsilon$ values, demonstrating consistent robustness and effectiveness across different settings.}
    
    \resizebox{ \columnwidth}{!}
    {\begin{tabular}{@{}cccccc} 
    
    \specialrule{1.5pt}{\aboverulesep}{\belowrulesep}
    \multirow{2}{*}{\textbf{Epsilon}} & \multirow{2}{*}{\textbf{Task}} & \multicolumn{4}{c}{\textbf{Dataset}} \\
      \cmidrule(lr){3-6}

\textbf{} && \textbf{MVTec AD}& \textbf{VisA}& \textbf{BTAD} & \textbf{BraTS2021}\\ 

    \specialrule{1.5pt}{\aboverulesep}{\belowrulesep} 

    \noalign{\vskip 1.5pt}

   \multirow{2}{*}{$\frac{2}{255}$} & AD & \textcolor{gray}{90.1 / }80.3 & \textcolor{gray}{89.6 / }77.8 & \textcolor{gray}{88.3 / }85.6 & \textcolor{gray}{95.7 / }86.3 \\

   \cmidrule(lr){3-6}

   & AL & \textcolor{gray}{94.0 / }81.6 & \textcolor{gray}{97.1 / }88.7 & \textcolor{gray}{93.7 / }79.1 & \textcolor{gray}{98.2 / }95.3\\

    \noalign{\vskip 1.5pt} 
    \specialrule{1pt}{\aboverulesep}{\belowrulesep}
    \noalign{\vskip 1.5pt} 

   \multirow{2}{*}{$\frac{4}{255}$} & AD & \textcolor{gray}{88.9 / }74.2 & \textcolor{gray}{88.8 / }75.4 & \textcolor{gray}{86.0 / }83.5 & \textcolor{gray}{94.8 / }83.4 \\

   \cmidrule(lr){3-6}

   & AL & \textcolor{gray}{93.2 / }76.4 & \textcolor{gray}{97.1 / }77.5 & \textcolor{gray}{93.1 / }75.4 & \textcolor{gray}{97.9 / }94.6\\

    \noalign{\vskip 1.5pt} 
    \specialrule{1pt}{\aboverulesep}{\belowrulesep}
    \noalign{\vskip 1.5pt} 

   \multirow{2}{*}{$\frac{8}{255} (Ours)$} & AD &  
   \textcolor{gray}{88.1 / }71.1 & \textcolor{gray}{88.5 / }74.3 & \textcolor{gray}{85.3 / }82.1 & \textcolor{gray}{94.3 / }81.0\\

   \cmidrule(lr){3-6}

   & AL & \textcolor{gray}{92.7 / }73.8 & \textcolor{gray}{96.9 /}85.2 & \textcolor{gray}{93.2  /}73.0 & \textcolor{gray}{97.7 /}94.5\\

   \noalign{\vskip 1.5pt} 
    \specialrule{1.5pt}{\aboverulesep}{\belowrulesep}
     \end{tabular}}
    \label{tab:diff_eps_eval}

\end{table}

In this section, we evaluate PatchGuard trained on PGD-10 with $l_{\infty}$ norm under $\epsilon = \frac{8}{255}$ using PGD-1000 with $l_2$ norms with various $\epsilon$. As shown in Table \ref{tab:diff_eps_eval}, our model remains robust against these types of attacks.

\begin{table}[h]
\caption{Evaluation of PatchGuard's robustness when trained on PGD-10 with $l_{\infty}$ norm under $\epsilon = \frac{8}{255}$, assessed using PGD-1000 with $l_2$ norms with various $\epsilon$. The results demonstrate the model's sustained robustness against different types of attacks.}    
    \resizebox{ \columnwidth}{!}
    {\begin{tabular}{@{}cccccc} 
    
    \specialrule{1.5pt}{\aboverulesep}{\belowrulesep}
    \multirow{2}{*}{\textbf{$\epsilon$}} & \multirow{2}{*}{\textbf{Task}} & \multicolumn{4}{c}{\textbf{Dataset}} \\
      \cmidrule(lr){3-6}

\textbf{} && \textbf{MVTec AD}& \textbf{VisA}& \textbf{BTAD} & \textbf{BraTS2021}\\ 

    \specialrule{1.5pt}{\aboverulesep}{\belowrulesep} 
    \noalign{\vskip 1.5pt} 
    
   \multirow{2}{*}{Clean} 
   & AD & 88.1 & 88.5 & 85.3 & 94.3\\
   \cmidrule(lr){3-6}
    & AL & 92.7 & 96.9 & 93.2 & 97.7  \\
    
   \noalign{\vskip 1.5pt} 
    \specialrule{1.5pt}{\aboverulesep}{\belowrulesep}

   \multirow{2}{*}{$\frac{16}{255}$} 
   & AD & 84.3 & 82.7 & 81.7 & 89.2\\
   \cmidrule(lr){3-6}
    & AL & 85.7 & 91.7 & 84.3 & 96.7\\
    
   \noalign{\vskip 1.5pt} 
    \specialrule{1.5pt}{\aboverulesep}{\belowrulesep}

   \multirow{2}{*}{$\frac{32}{255}$} 
   & AD & 82.1 & 81.0 & 79.5 & 87.4\\
   \cmidrule(lr){3-6}
    & AL & 83.6 & 90.3 & 82.9 & 96.1\\
    
   \noalign{\vskip 1.5pt} 
    \specialrule{1.5pt}{\aboverulesep}{\belowrulesep}

   \multirow{2}{*}{$\frac{64}{255}$} 
   & AD & 78.2 & 79.8 & 78.5 & 86.4\\
   \cmidrule(lr){3-6}
    & AL & 81.0 & 88.7 & 79.6 & 95.7\\
    
   \noalign{\vskip 1.5pt} 
    \specialrule{1.5pt}{\aboverulesep}{\belowrulesep}

   \multirow{2}{*}{$\frac{128}{255}$} 
   & AD & 77.2 & 78.6 & 76.7 & 84.7\\
   \cmidrule(lr){3-6}
    & AL & 78.3 & 86.7 & 77.9 & 95.0\\

   \noalign{\vskip 1.5pt} 
    \specialrule{1.5pt}{\aboverulesep}{\belowrulesep}

     \end{tabular}}
    \label{tab:l2_norm}

\end{table}

\section{Limitations} \label{appendix:limitations}

Our proposed PatchGuard method includes a ``foreground-aware anomaly generation" component that leverages Grad-CAM, which inherently ties our approach to a pretrained model. While this dependency enables our method to focus on relevant regions, it also introduces reliance on the quality and biases of the pretrained model. Furthermore, although we employ soft augmentations to encourage this component to identify accurate regions, there is no theoretical guarantee that it consistently achieves this objective. Nonetheless, as our empirical results demonstrate, the component performs well in practice, effectively highlighting anomalies in diverse scenarios.

\section{Trade-Off Between Anomaly Detection and Localization}\label{appendix:tradeoff-AD-AL}
  
The anomaly score and localization map of a method play a crucial role in shaping the design of attacks, enabling attackers to target either anomaly localization or detection with greater precision. In this study, however, we design our attacks on other methods to simultaneously target both localization and detection. In our proposed method, PatchGuard, the anomaly score is derived as the average of the top-k values in the anomaly map. This mechanism ensures that the optimal attack strategy for anomaly detection inherently aligns with the strategy for anomaly localization.

A particularly noteworthy aspect of our study is the approach we use to attack anomaly localization. Specifically, we flip anomaly patches to appear normal and normal patches to appear anomalous. An alternative logical attack could involve manipulating normal images to make all pixels anomalous, while for anomalous samples, the attack would preserve the normal pixels as they are and convert the anomalous pixels to appear normal. This approach would ultimately make the anomaly map of an anomalous sample indistinguishable from that of a normal sample.

Although this alternative attack is specifically designed for anomaly detection, it is far less effective for anomaly localization. Existing methods, even without explicitly addressing this type of targeted attack, are already highly vulnerable to detection-based attacks. In contrast, our method has been experimentally shown to be robust against such attacks. This robustness arises from our use of stronger adversarial training strategies, where all pixels are flipped to create more challenging adversarial examples during the training process.

\begin{figure*}[t]
  \begin{center}
    \includegraphics[width=.8\linewidth]{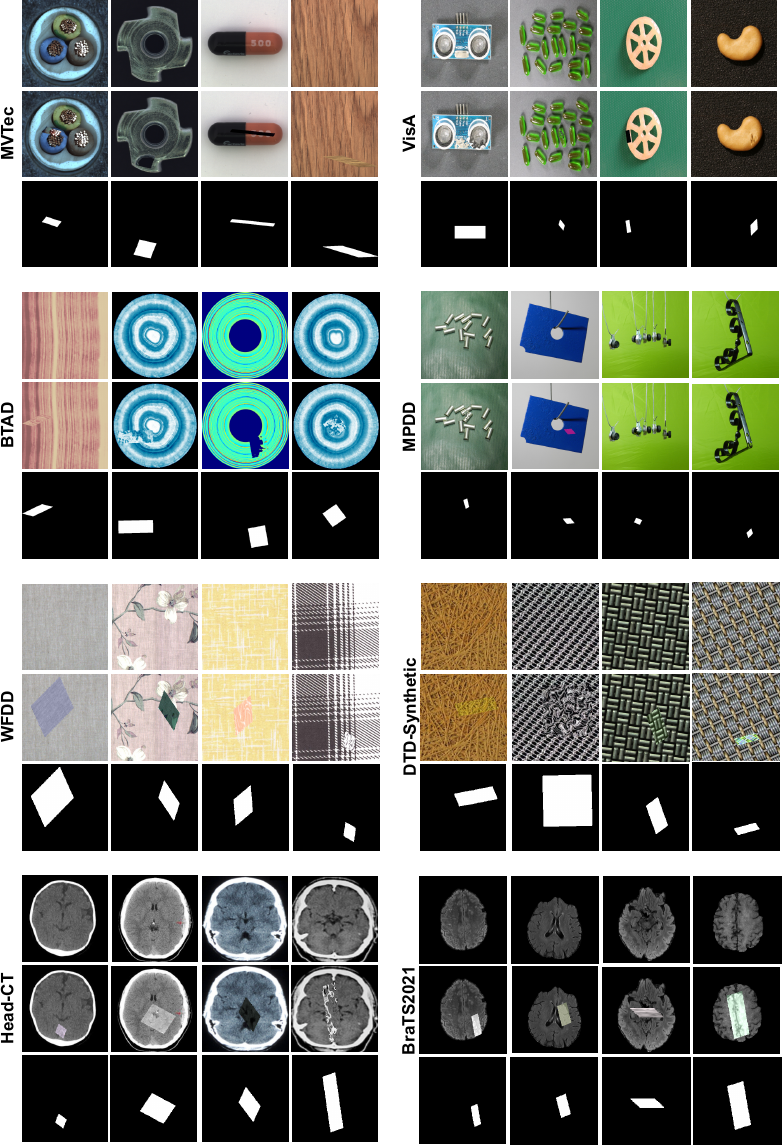}
    \caption{Visualization of Pseudo-Anomaly Generated for Each Dataset. Each group corresponds to one dataset: MVTec AD, VisA, BTAD, MPDD, WFDD, DTD-Synthetic, BraTS2021, and Head-CT. Within each group, columns represent randomly selected samples from the respective dataset. The first row shows a normal image, the second row depicts the corresponding pseudo-anomaly generated image, and the third row illustrates the associated anomaly mask.}

    \label{fig:aug figure}
  \end{center}
\end{figure*}

\end{document}